%% file: main.tex
\documentclass[]{alaya}
\usepackage{makecell}
\usepackage{wrapfig}
\usepackage{tabularx}
\usepackage{textcomp}
\usepackage{stfloats}
\usepackage{url}
\usepackage{verbatim}
\usepackage{titlesec}
\usepackage{adjustbox}
\usepackage{multirow}
\usepackage{pifont}
\usepackage[sc]{mathpazo}
\usepackage{comment}
\usepackage{amsmath,amssymb}
\usepackage{colortbl}
\usepackage{natbib}
\usepackage{color}
\usepackage{booktabs}
\usepackage{hyperref}
\usepackage{graphicx}
\usepackage{subcaption}
\usepackage{siunitx}
\usepackage[table]{xcolor}
\usepackage{caption}
\usepackage{float}
\usepackage{needspace}
\usepackage{xspace}

\makeatletter
\DeclareRobustCommand\onedot{\futurelet\@let@token\@onedot}
\def\@onedot{\ifx\@let@token.\else.\null\fi\xspace}

\makeatother

\newcommand{\yes}{\textcolor{green!60!black}{\checkmark}}
\newcommand{\no}{\textcolor{red!80!black}{$\times$}}

\definecolor{markgreen}{HTML}{009000}
\definecolor{markred}{HTML}{ea4335}

\newlength\savewidth

\renewcommand{\paragraph}[1]{\vspace{1.25mm}\noindent\textbf{#1}}

\title{InfiniDream: Open-Ended Photorealistic 3D Worlds Through Procedural Generation}

\author[1,2]{Yi-Chuan Huang}
\author[2]{Hao-Jen Chien}
\author[1,3]{Zheng-Hui Huang}
\author[2]{Yu-Lun Liu}
\author[1]{Kaipeng Zhang}
\author[1]{Zhixiang Wang}

\affiliation[1]{Alaya Lab}
\affiliation[2]{National Yang Ming Chiao Tung University}
\affiliation[3]{National Taiwan University}

\input{sec/00.abstract}

\date{October 20, 2026}

\begin{document}
\maketitle


\begin{figure}[H]
    \centering
    \includegraphics[width=\linewidth]{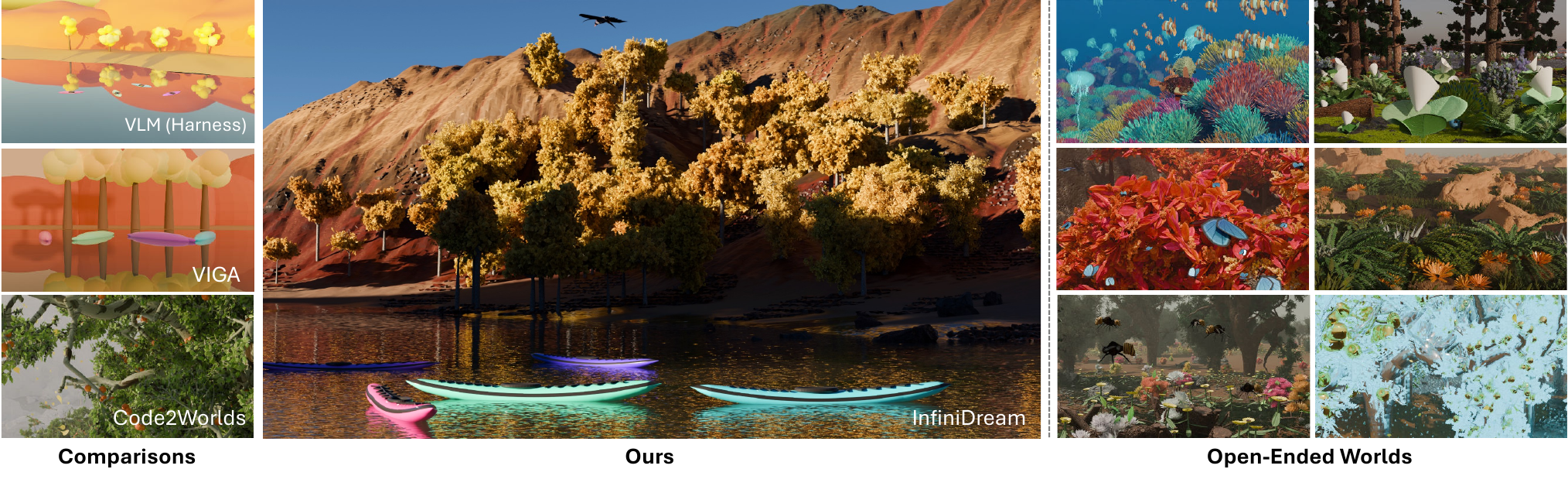}
    \captionof{figure}{
        \textbf{Beyond Predefined Factories: Rich, Editable Procedural Worlds.}
        InfiniDream extends an existing procedural generator to create novel assets and visually rich 3D scenes.
        We compare its results with existing VLM-based
        3D scene generation methods.
    }
    \label{fig:teaser}
\end{figure}

\input{sec/01.intro}
\input{sec/02.relatedwork}
\input{sec/03.method}
\input{sec/04.experiments}
\input{sec/05.conclusion}

\bibliographystyle{abbrv}
\bibliography{references}

\clearpage
\input{sec/06.appendix}

\end{document}

%% file: sec/00.abstract.tex
\abstract{
Procedural generation plays a key role in synthetic data creation and game development, offering greater controllability, editability, and diversity than static mesh representations. However, VLM-based code generation often produces relatively simple 3D assets, while sophisticated photorealistic procedural systems require extensive manual development. Existing VLM-guided procedural methods remain largely constrained by predefined factories and APIs, limiting their ability to create novel asset types.
We introduce \textbf{InfiniDream}, a VLM-driven framework for open-ended procedural 3D world generation. First, we propose \textit{Symbolically Constrained Asset Generation}, which leverages existing procedural algorithms to create novel, photorealistic assets while preserving editability and seed-based diversity. Second, we develop an end-to-end text-to-world pipeline that combines semantic scene planning with \textit{Deterministic Frustum-Guided Grid Placement} to construct large-scale 3D worlds with precise object placement, procedural population, and animation.
Extensive evaluations on 100 object and 20 scene generation tasks demonstrate improvements across four perceptual metrics. Diversity analysis, ablation studies, and human preference evaluations further validate the effectiveness of our framework. 

\par\medskip
\noindent
\textbf{Project page:} \url{https://yichuanh.github.io/infinidream-page/}\\
\textbf{Code:} \url{https://github.com/yichuanH/InfiniDream}\\
\noindent\textbf{Correspondence:}
\href{mailto:wangzx1994@gmail.com}{wangzx1994@gmail.com},
\href{mailto:kaipeng.zhang@shanda.com}{kaipeng.zhang@shanda.com},
\href{mailto:yulunliu@cs.nycu.edu.tw}{yulunliu@cs.nycu.edu.tw}

}

%% file: sec/01.intro.tex
\section{Introduction}

Procedural generation creates 3D assets and scenes
through executable programs, offering editability,
diversity, and scalability. These advantages have
made it widely applicable to synthetic data generation,
game development, and virtual world creation
\citep{infinigen2023infinite}. However, conventional
procedural generation methods rely heavily on
handcrafted rules and predefined asset generators,
limiting the variety of generated content and their
ability to create novel assets and scenes.

Recent advances in the code generation and tool-use
capabilities of large language models (LLMs) and
vision-language models (VLMs) have enabled the
generation and editing of 3D assets and scenes
through executable Blender code
\citep{hu2024scenecraft,lu2025ll3m,rondelli2026blenderrag,
yin2026viga,he2026thinking,wang2026threedcos}.
However, while existing approaches can construct
scenes by composing assets or directly synthesizing
geometry through code, generating large-scale 3D
scenes with consistently rich geometric details,
complex materials, and photorealistic appearance
remains challenging.

Infinigen~\citep{infinigen2023infinite} is built
on over 40,000 lines of code, including carefully
handcrafted procedural asset factories, to generate
photorealistic, editable 3D scenes. However, its
generative capabilities remain constrained by
predefined asset categories and procedural rules.

Recent studies have explored using LLMs and VLMs
to control sophisticated procedural generation
systems through function orchestration, parameter
adjustment, and the composition of existing generators
\citep{sun20233dgpt,zhou2025scenex,zhang2026code2worlds}.
While these approaches improve the controllability
of procedural scene generation, methods that primarily
operate within predefined functions and parameter
spaces remain limited in their ability to create
novel asset types beyond the capabilities of
existing generators.

Another line of research explores large-scale 3D
world generation using retrieved or generatively
synthesized mesh assets
\citep{guo2026worldclaw,wang2026worldgen}.
Although these methods support instance-level editing,
their mesh-based assets lack the parameterized
editability and seed-based diversity of procedural
generators. Producing diverse object variations
often requires direct mesh manipulation or regeneration.

\begin{figure*}[!t]
    \centering
    \includegraphics[
        width=\textwidth,
        keepaspectratio
    ]{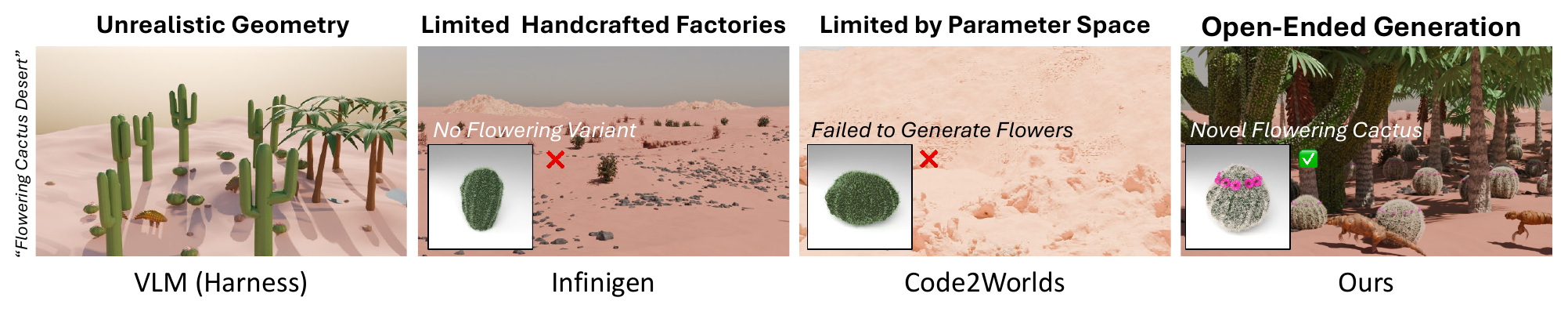}

    \caption{
        \textbf{Motivation: Beyond predefined procedural factories.}
        Given a flowering cactus desert prompt,
        VLM-based code generation produces simplified
        geometry, predefined factories offer limited
        variations, and parameter-based generation
        struggles to introduce new features.
        InfiniDream extends existing procedural
        generators to create novel assets within
        rich, photorealistic 3D worlds.
    }
    \label{fig:motivation}
\end{figure*}

Figure~\ref{fig:motivation} highlights three limitations:
VLM Harness produces less photorealistic scenes,
Infinigen is restricted to predefined factories,
and Code2Worlds~\citep{zhang2026code2worlds},
a recent VLM-guided Infinigen framework,
is limited to function calls and parameter adjustment.
These limitations motivate open-ended procedural
generation beyond existing factory capabilities.

\begin{wrapfigure}[10]{r}{0.40\columnwidth}
    \centering
    \vspace{-16pt}
    \includegraphics[
        width=\linewidth
    ]{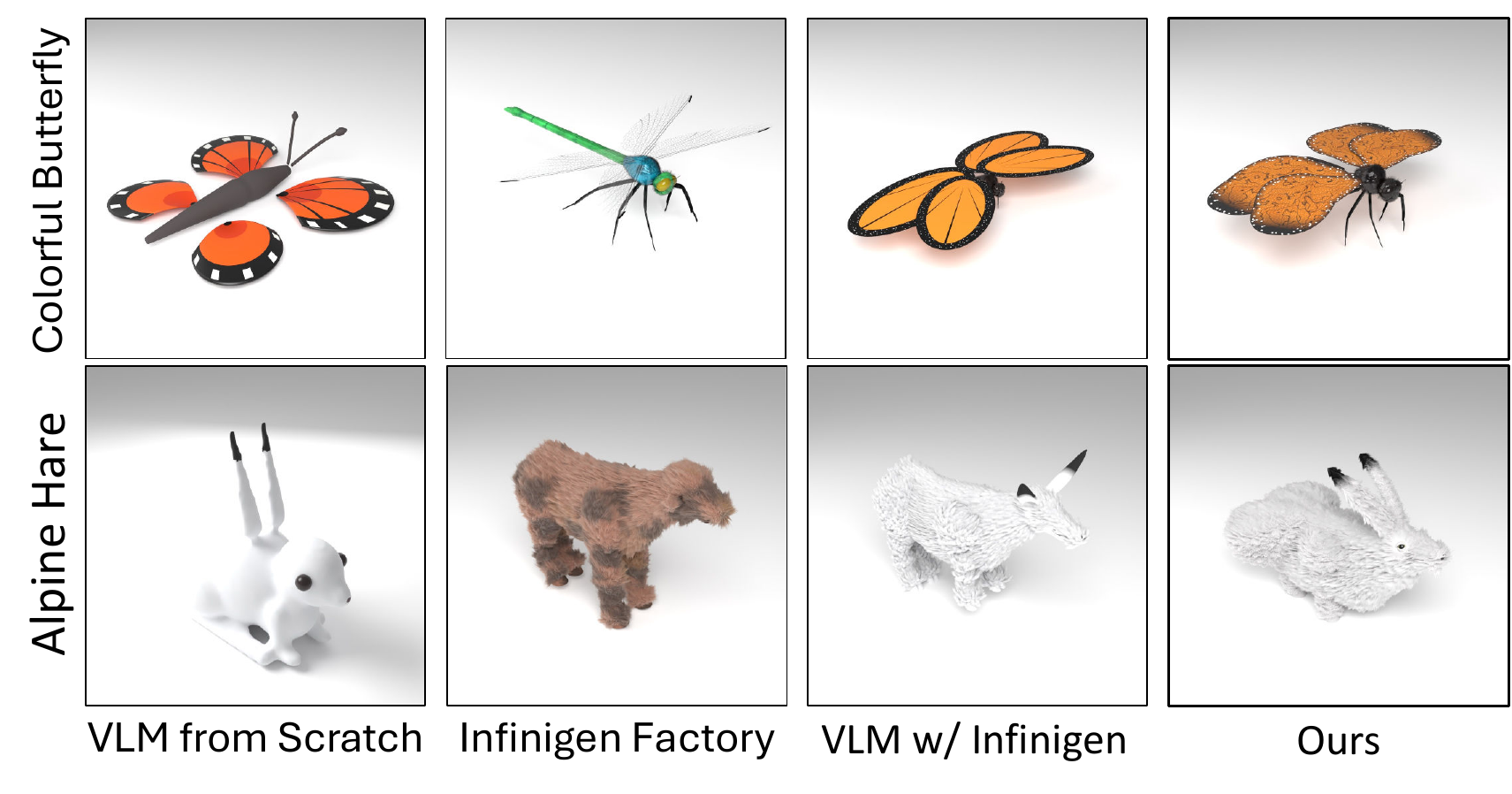}

    \caption{
        \textbf{Comparison of procedural asset generation strategies.}
        From left to right: VLM-generated Blender code
        from scratch, predefined Infinigen factories,
        VLM Agent Harness with full Infinigen access,
        and our symbolically constrained editing.
    }

    \label{fig:symbolic_editing}
    \vspace{-10pt}
\end{wrapfigure}

To address these challenges, we introduce
\textbf{InfiniDream}, a VLM-driven framework for
open-ended procedural 3D world generation,
designed to achieve realism, open-endedness,
editability, diversity, and scalability.
By employing VLM reasoning to compose and extend
sophisticated procedural programs, InfiniDream
generates novel assets beyond predefined factories
while leveraging existing procedural algorithms.
Our framework enables the creation of photorealistic,
editable, and diverse large-scale 3D worlds from
natural language descriptions.

At the asset level, Fig.~\ref{fig:symbolic_editing}
further motivates our Symbolically Constrained Editing
(SCE) strategy, which combines the flexibility of
VLM-based code generation with the rich procedural
structures of existing factories.

Our main contributions are summarized as follows:

\begin{enumerate}

    \item \textbf{Open-Ended Procedural Asset Generation.}
    We propose Symbolically Constrained Editing (SCE)
    to generate novel, photorealistic procedural assets
    with rich geometric details beyond predefined
    factories, while preserving editability and
    seed-based diversity.

    \item \textbf{Scalable Procedural World Generation.}
    We introduce an end-to-end text-to-world framework
    combining semantic scene planning and deterministic
    frustum-guided grid placement to generate large-scale,
    editable 3D worlds.

    \item \textbf{Comprehensive Experimental Validation.}
    We conduct extensive object- and scene-level
    evaluations, diversity analysis, ablation studies,
    and human preference studies to validate the
    effectiveness of InfiniDream.

\end{enumerate}

%% file: sec/02.relatedwork.tex
\section{Related Work}

\paragraph{Procedural 3D Asset Generation.}
Procedural modeling encodes 3D content as executable rules, from L-systems~\citep{lindenmayer1968lsystems,prusinkiewicz1990plants} and shape grammars~\citep{wonka2003instant,muller2006buildings} to learned procedural shape and CAD programs~\citep{sharma2018csgnet,tian2019shapeprograms,jones2020shapeassembly,jones2021shapemod,jones2023shapecoder,ritchie2023neurosymbolic,wu2021deepcad,khan2024text2cad}.
Advances in code generation~\citep{chen2021codex} and tool use~\citep{schick2023toolformer,yao2023react} now let language models synthesize editable 3D assets as Blender code~\citep{du2024blenderllm,lu2025ll3m,rondelli2026blenderrag,wang2026threedcos,yamada2025l3go,dai2025meshcoder,raistrick2026procfunc}, other shape programs~\citep{zhang2025shapecraft,jones2026shapelib,wang2026adsl,lin2026procedura}, or CAD scripts for mechanical parts~\citep{alrashedy2025cadcodeverify,wang2025cadfusion,rukhovich2025cadrecode,mallis2025cadassistant}.
Multimodal reasoning, retrieval augmentation~\citep{lewis2020rag}, and iterative visual feedback~\citep{alrashedy2025cadcodeverify,wang2025cadfusion,wang2026adsl,lin2026procedura} further improve reconstruction, editing, and visual fidelity~\citep{huang2024blenderalchemy,yin2026viga,he2026thinking,wang2026threedcos}.
Procedural material graphs are likewise generated or inferred from images~\citep{shi2020match,guerrero2022matformer,hu2023procmaterial,li2024procmaterialrl,li2025vlmaterial}, whereas inverse procedural modeling fits predefined procedural generators to targets~\citep{talton2011metropolis,vanegas2012inverse,stava2014trees,pearl2025geocode,zhao2025dipcg} and remains bounded by their categories.
Still, reliably synthesizing complex procedural geometry with rich geometric details and photorealistic appearance remains challenging.
Benchmarks for 3D code generation~\citep{gao2026threedcodebench,wang2026threedcos,du2024blenderllm,alrashedy2025cadcodeverify,gu2025blendergym,yang2026p3dbench,liu2026threedharnessbench} assess executability, geometric fidelity, and the benefits of execution feedback~\citep{chen2024selfdebug} and agentic coding~\citep{yang2024sweagent}.
However, 3DCodeBench~\citep{gao2026threedcodebench} draws its procedural modeling categories from predefined Infinigen factories~\citep{infinigen2023infinite,raistrick2024infinigenindoors}, leaving procedural asset generation beyond these categories underexplored.

\paragraph{3D World Generation.}
Procedural content generation~\citep{smelik2014survey,hendrikx2013pcg,shaker2016pcgbook} has long produced noise-based terrains~\citep{perlin1985noise,musgrave1989terrain} and cities~\citep{parish2001cities}, and now yields photorealistic natural and indoor scenes~\citep{infinigen2023infinite,raistrick2024infinigenindoors,raistrick2026procfunc}, simulation-ready assets~\citep{joshi2026infinigenarticulated}, and houses for embodied AI~\citep{deitke2022procthor}.
Language models have been leveraged for 3D scene planning, spatial reasoning, and executable scene construction~\citep{hu2024scenecraft,sun20233dgpt,zhou2025scenex,zhang2026code2worlds,feng2023layoutgpt,yang2024holodeck,celen2024idesign,fu2024anyhome,sun2025layoutvlm,aguinakang2024openuniverse,zhang2025scenelanguage,tam2025scenemotifcoder,gumin2025procscene}, and one line couples them with existing procedural generators through function orchestration, parameter adjustment, and composition of predefined procedural assets~\citep{sun20233dgpt,zhou2025scenex,qian2024scenecopilot,zhang2026code2worlds,zhang2024cityx,liu2025worldcraft,tang2025unrealllm,liu2026citygenagent}.
Related approaches~\citep{hu2024scenecraft,yin2026viga,zhang2026code2worlds} also explore hierarchical generation~\citep{fu2024anyhome,sun2025hierarchically,wang2024architect}, iterative visual feedback~\citep{ling2026scenethesis}, physics-aware dynamic simulation~\citep{lv2024gpt4motion}, and looping animation of static 3D scenes~\citep{xie2026ouroworld}.
However, methods restricted to orchestrating predefined procedural generators remain confined to their asset categories and parameter spaces, while semantically consistent and geometrically precise object placement in complex scenes remains challenging.
Another line of work builds large-scale worlds~\citep{guo2026worldclaw,wang2026worldgen} from procedural layouts, retrieved assets~\citep{deitke2023objaverse,yang2024holodeck,aguinakang2024openuniverse,deng2024citycraft}, generative 3D models~\citep{poole2023dreamfusion,hong2024lrm,zhang2024clay,xiang2025trellis}, and object-level composition~\citep{wang2024architect,huang2025midi,meng2026scenegen}, expanding generatable content and enabling instance-level manipulation.
However, these approaches and neural scene generators~\citep{hollein2023text2room,li2024genrc,fridman2023scenescape,yu2024wonderjourney,yu2025wonderworld,hunyuan2025hunyuanworld,huang2026vlm,chen2026pantheon360,cohenbar2023setthescene,gao2024graphdreamer,zhou2024gala3d,engstler2025syncity,chen2023scenedreamer,lin2023infinicity,xie2024citydreamer,lee2026skyfall,wu2024blockfusion,meng2025lt3sd} generally lack the reusable procedural structures and high-level parameters needed for procedural editing and seed-based geometric variation.
Orthogonally, recent work decouples explicit world states from generative video rendering~\citep{huang2026programmable,huang2026generative,liang2025diffusionrenderer} and benchmarks whether video models render what a program specifies~\citep{huang2026prowbench}; such systems still require an explicit world specification, which procedural world generation can supply.
InfiniDream addresses these gaps: it enables open-ended procedural asset generation beyond predefined factories and integrates it into scalable, fully procedural 3D world generation, while preserving procedural editability and seed-based diversity.

%% file: sec/03.method.tex
\section{Method}

We introduce \textbf{InfiniDream}, a VLM-driven framework for open-ended procedural 3D world generation. Given a text prompt, our framework selects a suitable initial scene and employs a Director to produce a structured Scene Plan specifying asset requirements, semantic placement constraints, and motion. Our framework consists of two core components: \textit{Symbolically Constrained Asset Generation}, which creates novel procedural assets through constrained editing, and \textit{Deterministic Frustum-Guided Grid Placement}, which translates semantic constraints into precise 3D object placements. Finally, procedural asset pooling, terrain refinement, and animation enable diverse, large-scale world generation.


\subsection{Symbolically Constrained Asset Generation}
\label{sec:sce}

\begin{figure}[t]
    \centering
    \includegraphics[width=\linewidth]{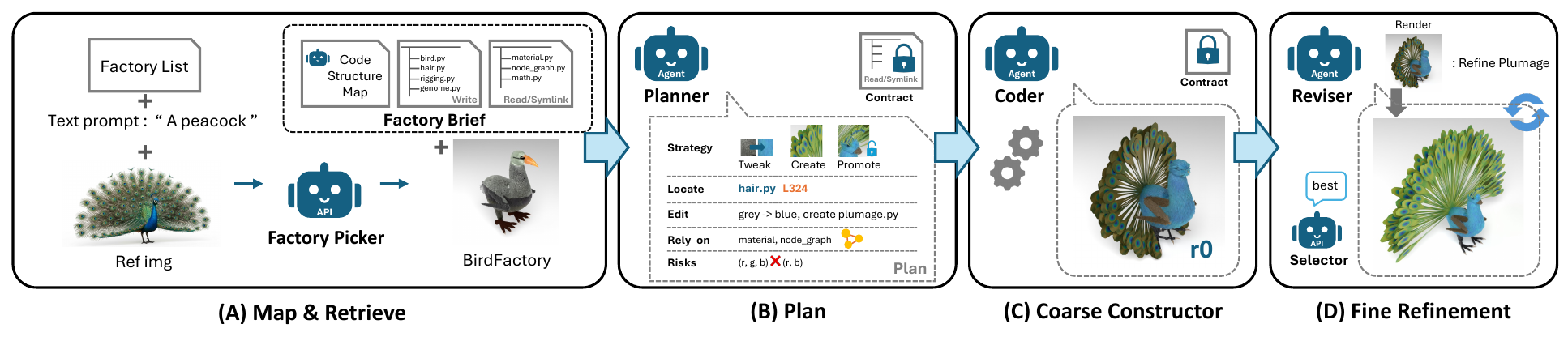}
    \caption{
        \textbf{Procedural asset generation through SCE.}
        (A) Factory mapping and retrieval.
        (B) Structured editing plan.
        (C) Coarse asset construction.
        (D) Render-guided fine refinement.
        Existing procedural algorithms are reused
        under explicit editing constraints.
    }
    \label{fig:pipeline_asset}
\end{figure}

Since directly providing a VLM with the entire Infinigen codebase often leads to imprecise editing (Fig.~\ref{fig:symbolic_editing}), we propose \textbf{Symbolically Constrained Editing (SCE)}, which extracts key code structures and explicit editing constraints, transforming a large procedural codebase into a manageable editing space (Fig.~\ref{fig:pipeline_asset}).

\subsubsection{Factory Mapping and Retrieval}

Given a text description and an optional reference image, the system retrieves the most relevant Infinigen factory and constructs a \textit{Factory Brief} (Fig.~\ref{fig:pipeline_asset}(A)). It contains a \textit{Code Structure Map} identifying key functions and dependencies, an \textit{Editable Surface} consisting of asset-specific modules dynamically identified through depth-limited dependency analysis, and \textit{Linked Infrastructure} providing read-only access to shared low-level algorithms (e.g., node graphs) via symbolic links. Together, these components establish explicit code-level constraints that restrict editing while preserving the underlying procedural algorithms. Further implementation details are provided in Appendix~\ref{app:sce}.

\subsubsection{Structured Planning and Coarse Construction}

The Planner produces a structured \textit{Editing Plan} specifying the editing strategy (\textit{Strategy}), target locations (\textit{Locate}), required modifications (\textit{Edit}), dependencies (\textit{Rely-on}), and potential risks (\textit{Risks}), as shown in Fig.~\ref{fig:pipeline_asset}(B). The plan also accounts for seed-dependent logic to preserve controllable procedural variation.

We define three editing strategies: \textit{Tweak} adjusts existing parameters and high-level logic; \textit{Create} introduces new procedural functions; and \textit{Promote} selectively extends editing permissions to protected modules when necessary. These strategies enable controlled expansion of the editing space.

The Coarse Constructor executes the Editing Plan under these constraints (Fig.~\ref{fig:pipeline_asset}(C)), preserving the original factory entry points and function signatures to maintain compatibility with the population pipeline and random seed control. It reuses existing algorithms to construct a coarse procedural asset. SCE preserves seed-dependent randomness by adjusting sampling ranges rather than fixing parameters to match the reference image.

\subsubsection{Fine Refinement}

As shown in Fig.~\ref{fig:pipeline_asset}(D), the Reviser iteratively compares rendered results with the target description and reference image, identifying discrepancies in geometry, materials, and visual details. It then performs targeted code modifications within the established editing boundaries. A Selector identifies the best candidate across refinement iterations.

Through constrained editing and iterative refinement, SCE generates novel procedural assets while preserving procedural editability and seed-controlled diversity.

\begin{figure*}[t]
    \centering
    \includegraphics[width=\textwidth]{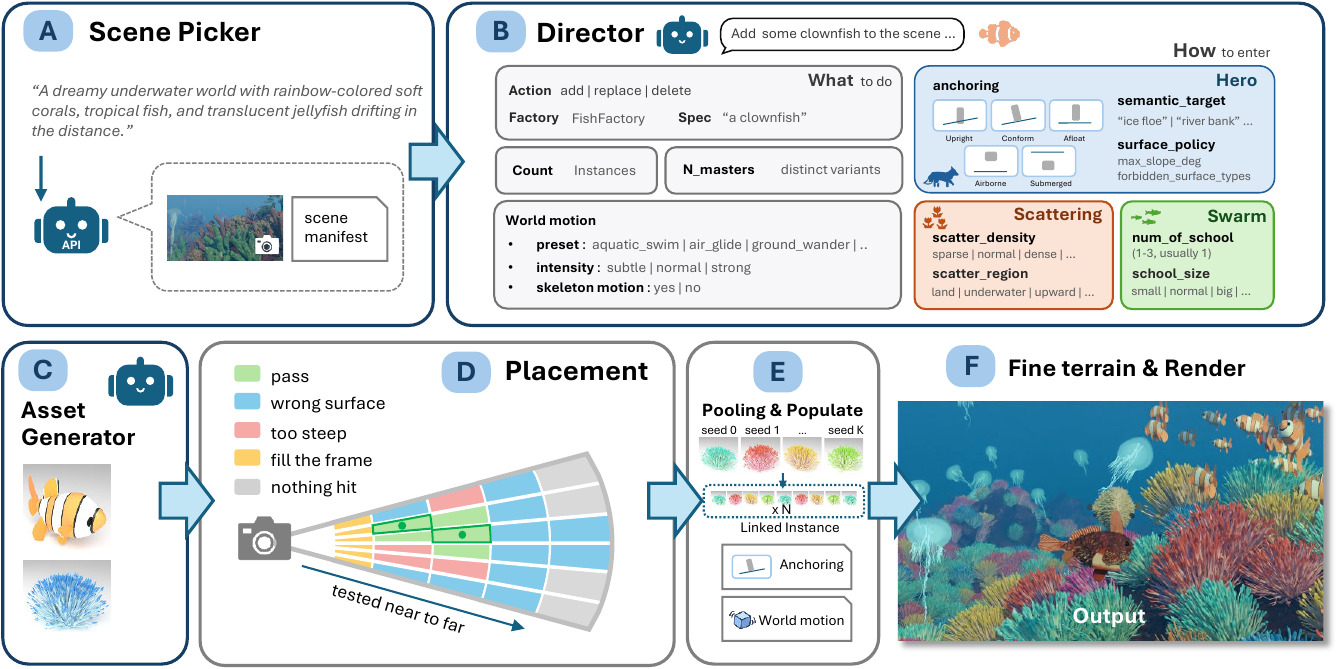}
    \caption{
        \textbf{Overview of InfiniDream's end-to-end world generation pipeline.}
        (A) Scene Picker collects the user instruction
        and current scene information.
        (B) Director generates a structured Scene Plan.
        (C) Asset Generator creates procedural assets.
        (D) Deterministic frustum-guided placement
        identifies valid object locations.
        (E) Population instantiates diverse assets.
        (F) Terrain refinement and rendering
        produce the final scene.
    }
    \label{fig:pipeline_e2e}
\end{figure*}


\subsection{Semantic Placement Reasoning}

Given a text prompt and current scene information, the
Director first formulates a natural-language plan and
converts it into a structured \textit{Scene Plan} specifying
editing operations (\texttt{add}, \texttt{replace}, and
\texttt{delete}), asset requirements, and semantic placement
constraints (Fig.~\ref{fig:pipeline_e2e}(A--B)).

The Director selects among three placement modes:
\textit{Hero} for semantically significant objects,
\textit{Scattering} for large-area distributions, and
\textit{Swarm} for collective entities. Hero placement
supports seven \textit{Anchoring} types (\texttt{rooted},
\texttt{upright}, \texttt{tilted}, \texttt{conform},
\texttt{afloat}, \texttt{airborne}, and \texttt{submerged}).
A \textit{Semantic Target} and \textit{Surface Policy}
further specify target regions, permissible surface types,
slope limits, and support conditions. Scattering and
Swarm are controlled by density, coverage, and group size.

The Director also determines the instance count $N$,
distinct procedural prototypes $K$, \textit{World Motion},
and \textit{Local Animation}. World Motion specifies
object trajectories and swept envelopes for collision
checking. VLM decisions are restricted to discrete
semantic operations and constraints, leaving continuous
geometric computations to deterministic placement.


\subsection{Deterministic Frustum-Guided Grid Placement}
\label{sec:placement}

To avoid repeated VLM-based coordinate prediction and costly render-based verification, we propose Deterministic Frustum-Guided Grid Placement, which translates the Director's semantic constraints into candidate 3D positions (Fig.~\ref{fig:pipeline_e2e}(D)). Candidates are evaluated in a near-to-far order without additional VLM calls or iterative rendering.

\subsubsection{Frustum-based Candidate Generation}

Given the specified placement region, we discretize the camera's image plane into a 2D grid and generate 3D candidate positions along the corresponding camera rays. For surface-supported objects, candidates are determined by ray intersections with scene geometry. For airborne and submerged objects, candidates are sampled at discrete depth intervals along each ray.

\subsubsection{Constraint-based Filtering and Depth-aware Placement}

Each candidate undergoes five deterministic checks:
(1) \textit{Visibility}, (2) \textit{Collision} with motion
swept envelopes, (3) \textit{Inter-instance Spacing},
(4) \textit{Surface Policy}, including surface type,
slope limits, and full support, and (5) \textit{Projected-area Band},
requiring $a_{\min} \leq a(p) \leq a_{\max}$.

Given depth-sorted candidates $\mathcal{C}=\{p_1,\ldots,p_n\}$,
where $g_j(p)\in\{0,1\}$ indicates satisfaction of constraint $j$,
we select the nearest valid candidate:
\begin{equation}
    i^* = \min \left\{
        i \;\middle|\;
        \bigwedge_{j=1}^{5} g_j(p_i) = 1
    \right\},
    \qquad
    p^* = p_{i^*}.
    \label{eq:valid_placement}
\end{equation}

Combined with projected-area bounds, near-to-far search
enables implicit size-aware placement, favoring closer
positions for smaller objects and farther positions
for larger ones. For multiple instances, the initial
search depth increases progressively, while previously
placed objects are included in subsequent collision checks.
If no valid position exists, hierarchical constraint
relaxation is applied (Appendix~\ref{app:placement}).


\subsection{Scalable Population and Animation}

We combine the procedural assets generated in Sec.~\ref{sec:sce} with the validated placements from Sec.~\ref{sec:placement} through semantically controlled asset pooling, terrain refinement, and procedural animation (Fig.~\ref{fig:pipeline_e2e}(E)).

\paragraph{Semantically Budgeted Asset Pooling.}
To reduce the repeated construction cost of complex procedural assets, we extend Infinigen's asset pooling mechanism to independent object population. The Director determines the number of instances $N$ and distinct procedural prototypes $K$ based on scene requirements. For assets supporting seed-based variation, we generate $K$ prototypes using different factory seeds and create $N$ instances through linked duplication:
\begin{equation}
    \mathcal{M} = \{M_k = G(s_k)\}_{k=1}^{K},
    \qquad
    I_i = \operatorname{Clone}
    \left(M_{((i-1)\bmod K)+1}\right).
    \label{eq:asset_pooling}
\end{equation}

Here, $G$ denotes the procedural generator, $s_k$ is the seed for the $k$-th prototype, and $I_i$ is the $i$-th instance. This mechanism reduces expensive asset constructions from $N$ to $K$ while maintaining seed-based diversity across prototypes. Assets unsuitable for pooling are generated independently.

\paragraph{Population, Terrain Refinement, and Animation.}
Procedural assets are placed according to their Anchoring modes
and integrated with Infinigen's fine terrain pipeline.
Animation combines \textit{World Motion} and \textit{Local Animation},
with shared motion specifications ensuring consistency
between placement constraints and actual trajectories.
The final scene is then refined and rendered (Fig.~\ref{fig:pipeline_e2e}(E--F)).

%% file: sec/04.experiments.tex
\section{Experiments}

We evaluate InfiniDream at both the object and scene levels, complemented by diversity experiments, ablation studies, and a user study.

\subsection{Evaluation Protocol and Baselines}
\label{sec:exp_setup}

\paragraph{Object-level Evaluation.}
We compare InfiniDream against five baselines: VLM (one-shot), VLM (harness), VIGA~\citep{yin2026viga}, BlenderRAG~\citep{rondelli2026blenderrag}, and Infinigen~\citep{infinigen2023infinite}. Following 3DCodeBench~\citep{gao2026threedcodebench}, we adopt its codebase for the two VLM baselines, evaluating single-turn code generation and iterative refinement with a native coding-agent harness. The remaining baselines represent visual-feedback refinement, retrieval-augmented generation, and original Infinigen factory outputs, respectively.

\paragraph{Scene-level Evaluation.}
We compare large-scale world generation against VLM (harness), VIGA~\citep{yin2026viga}, Infinigen~\citep{infinigen2023infinite}, and Code2Worlds~\citep{zhang2026code2worlds}, representing agentic code generation, visual-feedback refinement, original procedural scene generation, and LLM-driven hierarchical procedural orchestration, respectively.

\paragraph{Implementation and Metrics.}
We use Claude Opus 4.8 for InfiniDream and all VLM-based baselines, with Blender 4.x for generation and rendering. We evaluate 100 object-level tasks across six categories and 20 scene-level tasks at resolutions of $1024 \times 1024$ and $960 \times 540$, respectively.

We report CLIP~\citep{radford2021learningtransferablevisualmodels}, SigLIP2~\citep{tschannen2025siglip2multilingualvisionlanguage}, DINOv3~\citep{simeoni2025dinov3}, and ImageReward~\citep{xu2023imagerewardlearningevaluatinghuman}. At the object level, CLIP, SigLIP2, and ImageReward are averaged over four fixed views ($45^\circ$, $135^\circ$, $225^\circ$, and $315^\circ$), while DINOv3 compares the front view against four held-out reference images, independently generated from the same text prompt, and reports the mean cosine similarity. We additionally report code executability. Scene-level metrics use each method's primary camera view. For scene-level DINOv3, each method's primary camera view is compared against a single per-prompt scene reference image, generated by Gemini 3 Pro Image from the same scene prompt as a photorealistic wide-angle photograph of the entire scene; the same reference is shared by all methods and is never provided to any method during generation. Further details are provided in the Appendix.

\paragraph{Diversity Evaluation.}
We compare InfiniDream with TRELLIS.2~\citep{xiang2025nativecompactstructuredlatents} on 15 text prompts. For each prompt, InfiniDream generates five variations using different random seeds, while TRELLIS.2 uses five independently generated reference images. We measure semantic consistency using SigLIP2 and visual diversity using average pairwise DINOv2 feature distance.

\subsection{Experimental Results}
\label{sec:exp_result}

\paragraph{Qualitative Comparisons}

\begin{wrapfigure}[9]{r}{0.45\textwidth}
    \vspace{-20pt}
    \centering
    \includegraphics[
        width=\linewidth,
        trim=0 15pt 0 0,
        clip
    ]{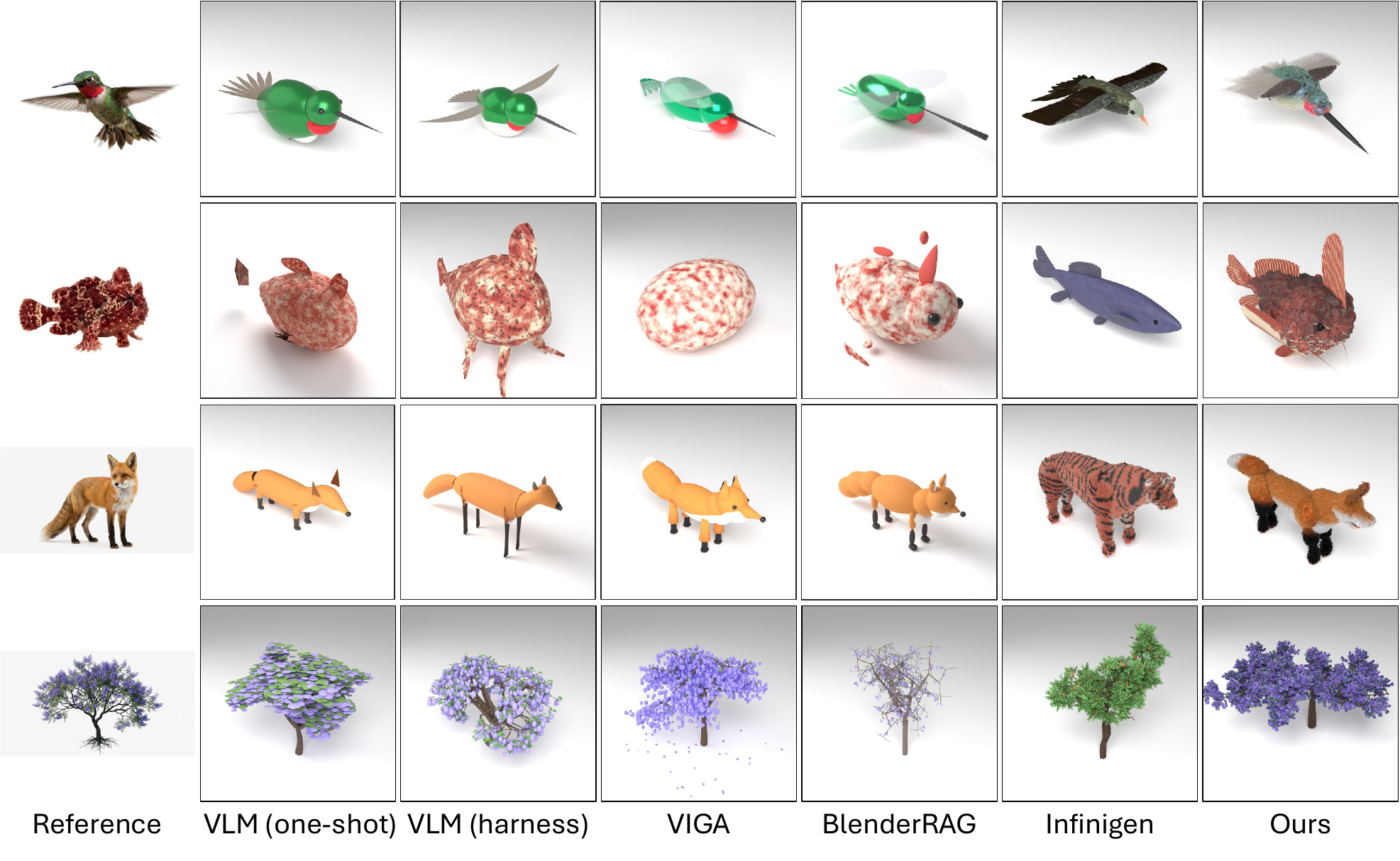}
    \par\vspace{3pt}
    \setlength{\abovecaptionskip}{0pt}
    \caption{
        \textbf{Qualitative comparison of
        object-level generation.}
        Generated 3D assets under identical
        text prompts.
    }
    \label{fig:object_comparison}
    \vspace{-10pt}
\end{wrapfigure}

\begin{figure}
    \centering
    \includegraphics[
        width=\linewidth
    ]{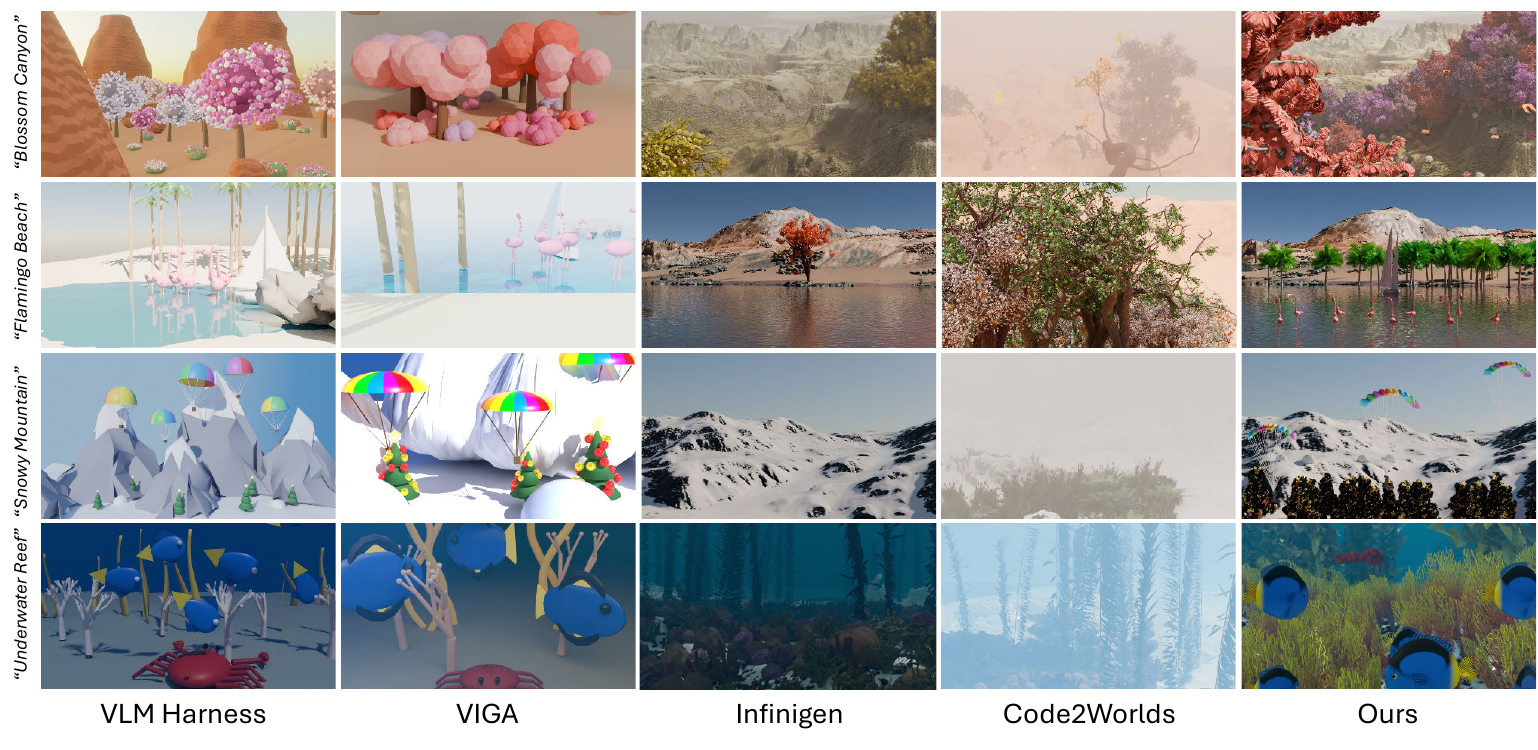}
    \caption{
    \textbf{From text to richly detailed 3D worlds.}
    Qualitative comparisons across four diverse environments,
    highlighting scene richness, visual fidelity, and semantic alignment.
    All methods use identical text prompts.
}
    \label{fig:scene_comparison}
\end{figure}

Figures~\ref{fig:object_comparison} and~\ref{fig:scene_comparison} present qualitative comparisons of InfiniDream with existing methods at the object and scene levels, highlighting differences in geometric detail, photorealism, and scene composition.

\paragraph{Quantitative Comparisons}

As shown in Tables~\ref{tab:object_results} and~\ref{tab:scene_results}, InfiniDream achieves the highest scores across all four perceptual metrics at both the object and scene levels.

\begin{table}[t]
    \centering

    \begin{minipage}[t]{0.50\linewidth}
        \centering
        \vspace{0pt}

    \caption{
        \textbf{Object-level comparison.}
        Results on 100 tasks across six categories.
    }
        \label{tab:object_results}

        \resizebox{\linewidth}{!}{%
        \begin{tabular}{lccccc}
            \toprule

            \textbf{Method}
            & \textbf{CLIP}$\uparrow$
            & \textbf{SigLIP2}$\uparrow$
            & \textbf{DINOv3}$\uparrow$
            & \textbf{IR}$\uparrow$
            & \textbf{Exec. (\%)}$\uparrow$ \\

            \midrule

            VLM (One-shot)
            & 27.21 & 0.120 & 0.419 & -1.17 & 85 \\

            VLM (Harness)
            & 29.20 & 0.139 & 0.498 & -0.82
            & \textbf{99} \\

            VIGA
            & 27.20 & 0.113 & 0.395 & -1.19 & 90 \\

            BlenderRAG
            & 27.56 & 0.127 & 0.406 & -1.15 & 91 \\

            Infinigen
            & 22.84 & 0.079 & 0.253 & -1.90 & -- \\

            \midrule

            \textbf{InfiniDream}
            & \textbf{30.07}
            & \textbf{0.148}
            & \textbf{0.554}
            & \textbf{-0.62}
            & 98 \\

            \bottomrule
        \end{tabular}%
        }
    \end{minipage}%
    \hfill%
    \begin{minipage}[t]{0.46\linewidth}
        \centering
        \vspace{0pt}

        \caption{
            \textbf{Scene-level comparison.}
            Results on 20 scene generation tasks.
        }
        \label{tab:scene_results}

        \resizebox{\linewidth}{!}{%
        \begin{tabular}{lcccc}
            \toprule

            \textbf{Method}
            & \textbf{CLIP}$\uparrow$
            & \textbf{SigLIP2}$\uparrow$
            & \textbf{DINOv3}$\uparrow$
            & \textbf{IR}$\uparrow$ \\
            \midrule

            VLM (Harness)
            & 28.85 & 0.126 & 0.321 & -0.894 \\

            VIGA
            & 27.14 & 0.121 & 0.246 & -1.433 \\

            Infinigen
            & 25.21 & 0.079 & 0.182 & -1.710 \\

            Code2Worlds
            & 22.76 & 0.072 & 0.114 & -2.014 \\

            \midrule

            \textbf{InfiniDream}
            & \textbf{29.64}
            & \textbf{0.143}
            & \textbf{0.361}
            & \textbf{-0.877} \\

            \bottomrule
        \end{tabular}%
        }
    \end{minipage}

\end{table}

\Needspace{13\baselineskip}
\paragraph{Human Preference Study}

\begin{wraptable}[12]{r}{0.35\columnwidth}
    \centering
    \vspace{-8pt}
    \fontsize{8}{9}\selectfont
    \setlength{\tabcolsep}{1pt}

    \caption{
        \textbf{Human preference study.}
        Preference rates for InfiniDream.
    }
    \label{tab:user_study}

    \begin{tabular*}{\linewidth}{@{\extracolsep{\fill}}lccc@{}}
        \toprule
        \textbf{Level}
        & \textbf{Align. (\%)}
        & \textbf{Quality (\%)}
        & \textbf{Overall (\%)} \\
        \midrule
        Object & 81.3 & 73.7 & \textbf{77.5} \\
        Scene  & 91.0 & 75.3 & \textbf{83.2} \\
        \bottomrule
    \end{tabular*}

    \vspace{10pt}

    \caption{
        \textbf{Diversity comparison.}
        Five variations per prompt.
    }
    \label{tab:diversity}

    \begin{tabular*}{\linewidth}{@{\extracolsep{\fill}}lcc@{}}
        \toprule
        \textbf{Method}
        & \textbf{SigLIP2}$\uparrow$
        & \shortstack{\textbf{DINOv2}\\
          \textbf{Dist.}$\uparrow$} \\
        \midrule
        TRELLIS.2 & \textbf{0.1317} & 0.6084 \\
        \textbf{InfiniDream} & 0.1225 & \textbf{0.6704} \\
        \bottomrule
    \end{tabular*}

    \vspace{-10pt}
\end{wraptable}

We conduct a human preference study with 30 participants across 10 object and 10 scene tasks. Participants compare results from competing methods based on semantic alignment and visual quality. Table~\ref{tab:user_study} reports InfiniDream's preference rates for each criterion and overall.

\subsection{Diversity Analysis}
\label{sec:diversity}

\begin{figure}[!t]
    \centering
    \includegraphics[
        width=\columnwidth
    ]{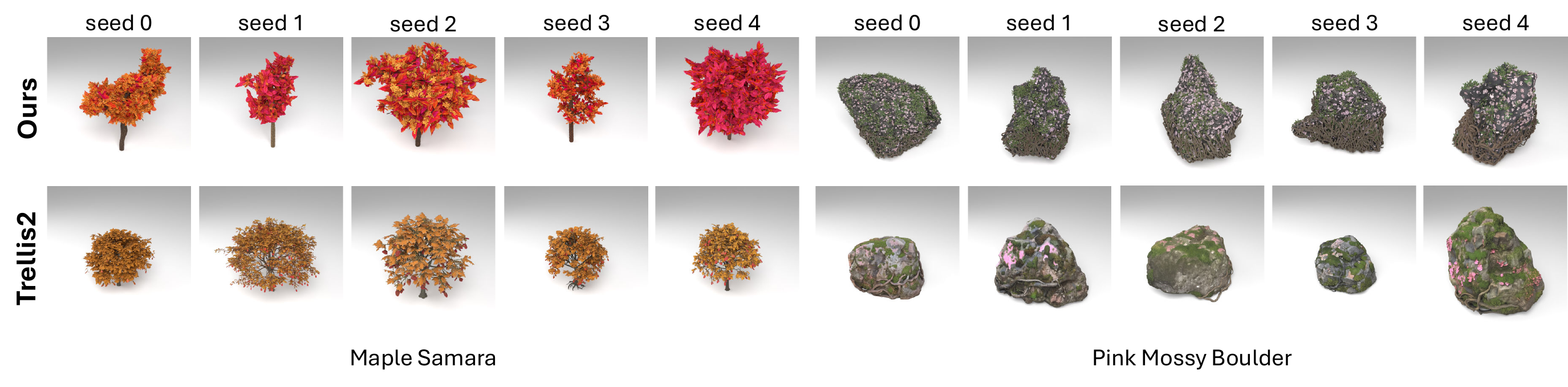}
    \vspace{-8pt}
    \caption{
        \textbf{High-quality and diverse procedural assets.}
        InfiniDream generates five variations by changing
        random seeds, while TRELLIS.2 uses five independently
        generated reference images.
    }
    \label{fig:diversity}
    \vspace{-10pt}
\end{figure}

For each text prompt, InfiniDream generates five
variations using different random seeds, while
TRELLIS.2 uses five independently generated reference
images. We evaluate semantic consistency using SigLIP2
and visual diversity using average pairwise DINOv2
feature distance. InfiniDream achieves greater visual
diversity while maintaining comparable semantic alignment
(Fig.~\ref{fig:diversity}, Table~\ref{tab:diversity}).

\subsection{Ablation Studies}

\paragraph{Object-level Ablation}

\begin{figure}[!t]
    \centering

    \begin{minipage}[t]{0.43\linewidth}
        \centering
        \captionsetup{type=table}

        \caption{
    \textbf{Quantitative object-level ablation.}
    Evaluation of generation components and
    refinement strategies across 25 animal and
    plant tasks, measuring semantic alignment,
    visual similarity, and perceptual quality.
}
        \label{tab:asset_ablation}

        \vspace{4pt}

        \resizebox{\linewidth}{!}{
        \begin{tabular}{lcccc}
            \toprule
            \textbf{Configuration}
            & \textbf{CLIP}$\uparrow$
            & \textbf{SigLIP2}$\uparrow$
            & \textbf{DINOv3}$\uparrow$
            & \textbf{IR}$\uparrow$ \\
            \midrule

            w/o Ref. Image
            & 28.68 & 0.1338 & 0.363 & -0.978 \\

            Random Factory
            & 27.39 & 0.1157 & 0.296 & -1.520 \\

            w/o Planner
            & 28.72 & 0.1274 & 0.368 & -1.023 \\

            Agent (Infinigen)
            & 29.01 & 0.1343 & 0.387 & -0.968 \\

            w/o Refiner
            & 28.80 & 0.1192 & 0.334 & -1.420 \\

            3 Rounds
            & 29.33 & 0.1354 & 0.438 & -0.824 \\

            \midrule

            \textbf{Full}
            & \textbf{30.37}
            & \textbf{0.1417}
            & \textbf{0.494}
            & \textbf{-0.580} \\

            \bottomrule
        \end{tabular}
        }
    \end{minipage}
    \hfill
    \begin{minipage}[t]{0.55\linewidth}
        \centering
        \vspace{0pt}

        \includegraphics[
            width=\linewidth
        ]{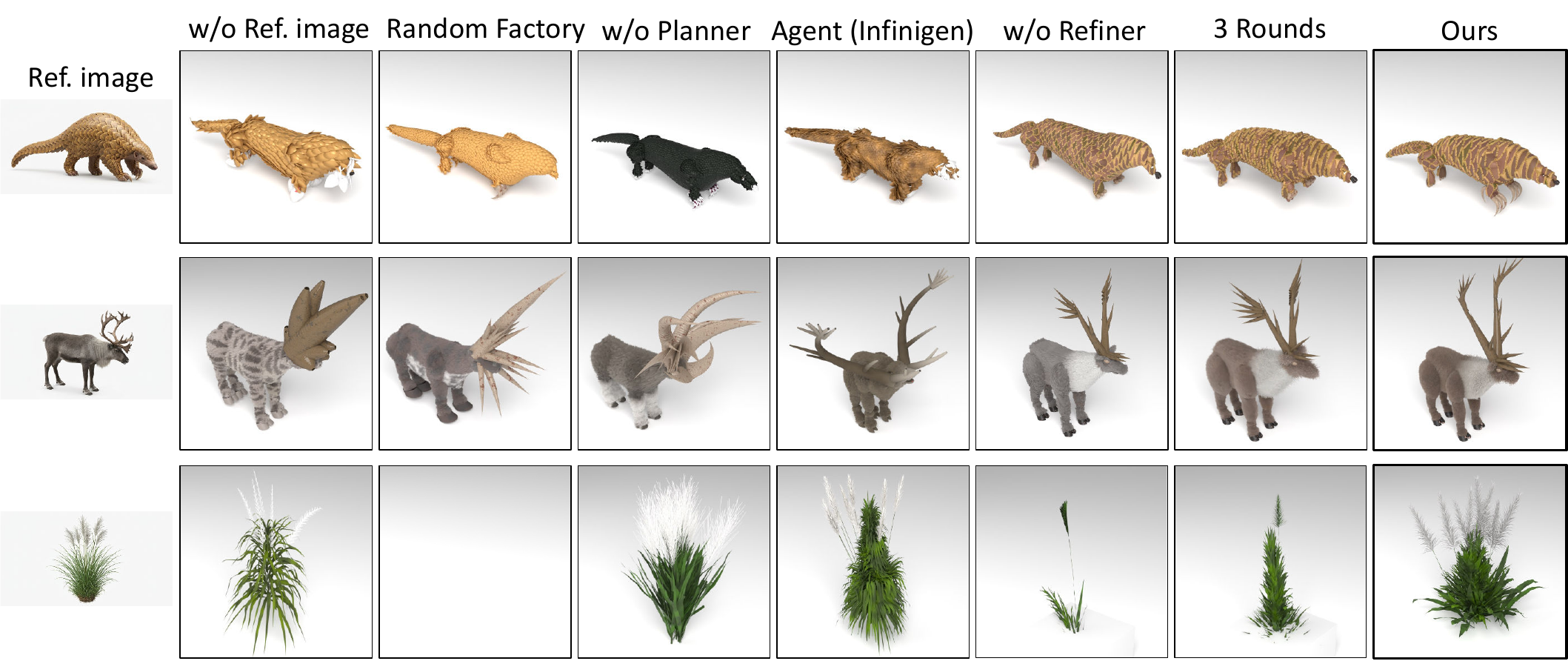}

        \caption{
    \textbf{Qualitative object-level ablation.}
    Visual comparison of different generation
    strategies, illustrating the effects of
    reference images, factory selection,
    structured planning, and iterative refinement
    on asset geometry, fine-grained details,
    and visual appearance.
}
        \label{fig:asset_ablation}
    \end{minipage}

\end{figure}

We evaluate six ablation settings on 25 animal and plant tasks: (1) \textit{w/o Ref. Image}, (2) \textit{Random Factory}, (3) \textit{w/o Planner}, (4) \textit{Agent (Infinigen)}, (5) \textit{w/o Refiner}, and (6) \textit{3 Rounds}. The full pipeline allows up to six refinement rounds. For Agent (Infinigen), Claude Opus 4.8 is given the entire Infinigen codebase and Bash access to autonomously write and edit code, render and inspect results, and determine when to stop within six rounds, without our structured planning or editing constraints. Quantitative and qualitative results are shown in Table~\ref{tab:asset_ablation} and Fig.~\ref{fig:asset_ablation}.

\paragraph{Scene-level Ablation.}
We evaluate four settings on 20 scenes:
(1) \textit{w/o Grid Search}, using direct 3D coordinate prediction;
(2) \textit{w/o Anchoring}, reverting to Infinigen's grounding;
(3) \textit{w/o Semantic Target}, removing semantic surface constraints;
and (4) \textit{Agent (Placement)}, allowing a VLM agent to autonomously place our generated 3D assets in the same selected coarse scene given the text prompt, without our placement pipeline.

\begin{figure*}[t]
    \centering

    \begin{minipage}[t]{0.35\textwidth}
        \vspace{0pt}
        \centering
        \captionof{table}{
    \textbf{Quantitative scene-level placement ablation.}
    Evaluation on 20 scenes measuring object
    readability, ground contact, and contact spread.
}
        \label{tab:scene_ablation}

        \resizebox{\linewidth}{!}{%
        \begin{tabular}{lccc}
            \toprule
            \textbf{Configuration}
            & \shortstack{\textbf{Unread.}\\(\%)$\downarrow$}
            & \shortstack{\textbf{Ground}\\P95 (m)$\downarrow$}
            & \shortstack{\textbf{Spread}\\P95 (m)$\downarrow$} \\
            \midrule
            w/o Grid Search
            & 27.4 & \textbf{0.15} & 0.25 \\

            w/o Anchoring
            & 40.3 & 0.48 & 0.74 \\

            w/o Semantic Target
            & 14.0 & \textbf{0.15} & 0.38 \\

            Agent (Placement)
            & 50.9 & 0.54 & 0.75 \\

            \midrule
            \textbf{Full (Ours)}
            & \textbf{7.8}
            & \textbf{0.15}
            & \textbf{0.24} \\
            \bottomrule
        \end{tabular}%
        }
    \end{minipage}
    \hfill
    \begin{minipage}[t]{0.62\textwidth}
        \vspace{0pt}
        \centering
        \includegraphics[
            width=\linewidth
        ]{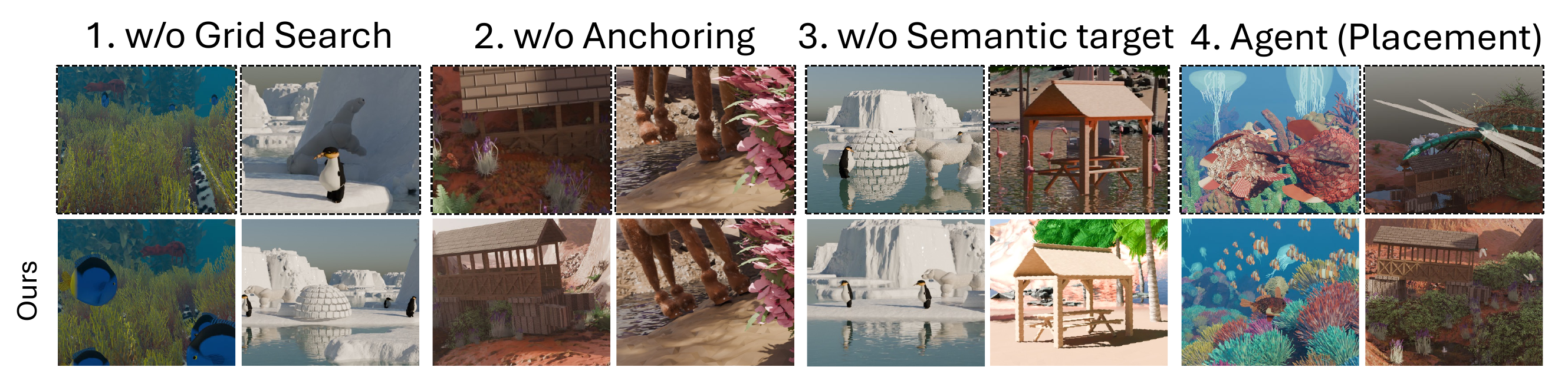}

    \captionof{figure}{
    \textbf{Qualitative scene-level placement ablation.}
    Visual comparison of different placement strategies,
    illustrating the effects of semantic planning,
    grid-based placement, and anchoring constraints
    on object visibility, spatial arrangement,
    and scene composition.
}
        \label{fig:scene_ablation}
    \end{minipage}

\end{figure*}

We evaluate three placement metrics:
\textit{Unreadable Rate}, the percentage of objects
outside the camera view or occupying less than $0.1\%$
of the image area;
\textit{Ground-contact Error}, the absolute distance between
an object's lowest contact point and its supporting surface,
capturing floating or penetration;
and \textit{Contact Spread}, the difference between
the maximum and minimum terrain-relative gaps
across sampled contact points, capturing uneven
grounding and partial suspension (surface-supported
objects only; see Appendix~\ref{app:ablation}).
We report the 95th percentile (P95) for the latter
two metrics. For Contact Spread, P95 is computed
over per-object values pooled across all
surface-supported requested-add objects in the 20 scenes.
Results are shown in Table~\ref{tab:scene_ablation}
and Fig.~\ref{fig:scene_ablation}.

%% file: sec/05.conclusion.tex
\section{Conclusion}

We presented \textbf{InfiniDream}, a VLM-driven framework for open-ended procedural 3D world generation. By combining Symbolically Constrained Asset Generation with semantic planning and deterministic placement, InfiniDream creates novel, photorealistic, and editable procedural assets and composes them into diverse, large-scale 3D worlds. Extensive experiments demonstrate its effectiveness at both the object and scene levels.

\paragraph{Limitations and Future Work.}
Our current animation pipeline relies on a predefined set of motion types. Future work could explore VLM-generated animation code to support more complex and customized motions. We also aim to extend InfiniDream to indoor environments with more complex spatial relationships and object interactions.




%% file: sec/06.appendix.tex
\beginappendix

\section{Additional Object-level Comparisons}
\label{app:object_comparison}

\begin{figure}[H]
    \centering
    \includegraphics[
        width=\textwidth,
        height=0.88\textheight,
        keepaspectratio
    ]{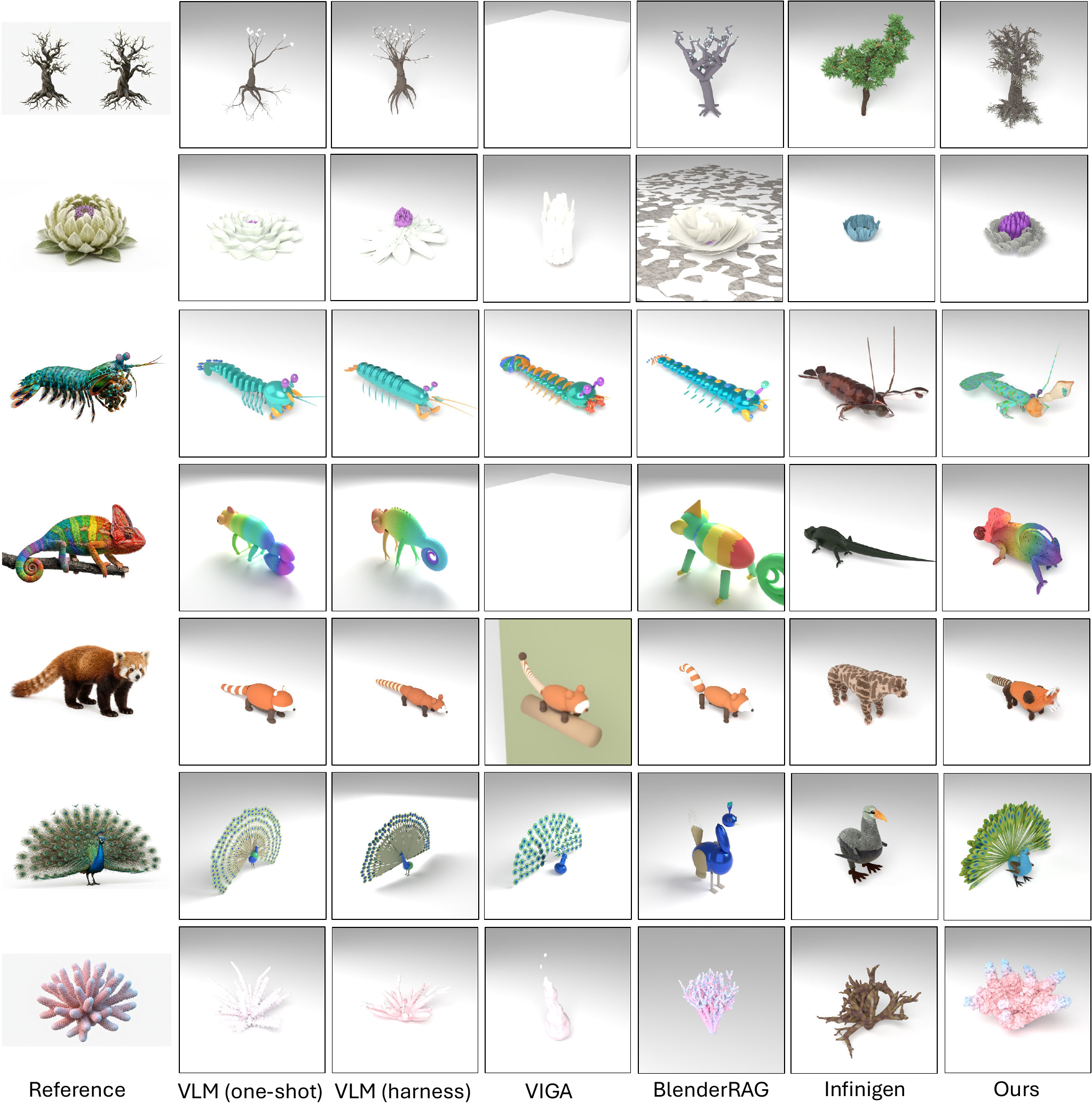}

    \caption{
        \textbf{Additional qualitative comparisons of object-level generation.}
        More examples comparing InfiniDream with existing
        methods under identical text prompts.
    }
    \label{fig:asset_app}
\end{figure}

We provide additional qualitative comparisons of object-level
generation across diverse asset categories in Fig.~\ref{fig:asset_app}.
All methods are evaluated using identical text prompts under
the same experimental protocol described in Sec.~\ref{sec:exp_setup}.
These additional examples further illustrate differences in
geometric complexity, material details, and visual fidelity
among the compared methods.

\section{Additional Scene-level Comparisons}
\label{app:scene_comparison}

\begin{figure}[H]
    \centering
    \includegraphics[
        width=\textwidth,
        height=0.88\textheight,
        keepaspectratio
    ]{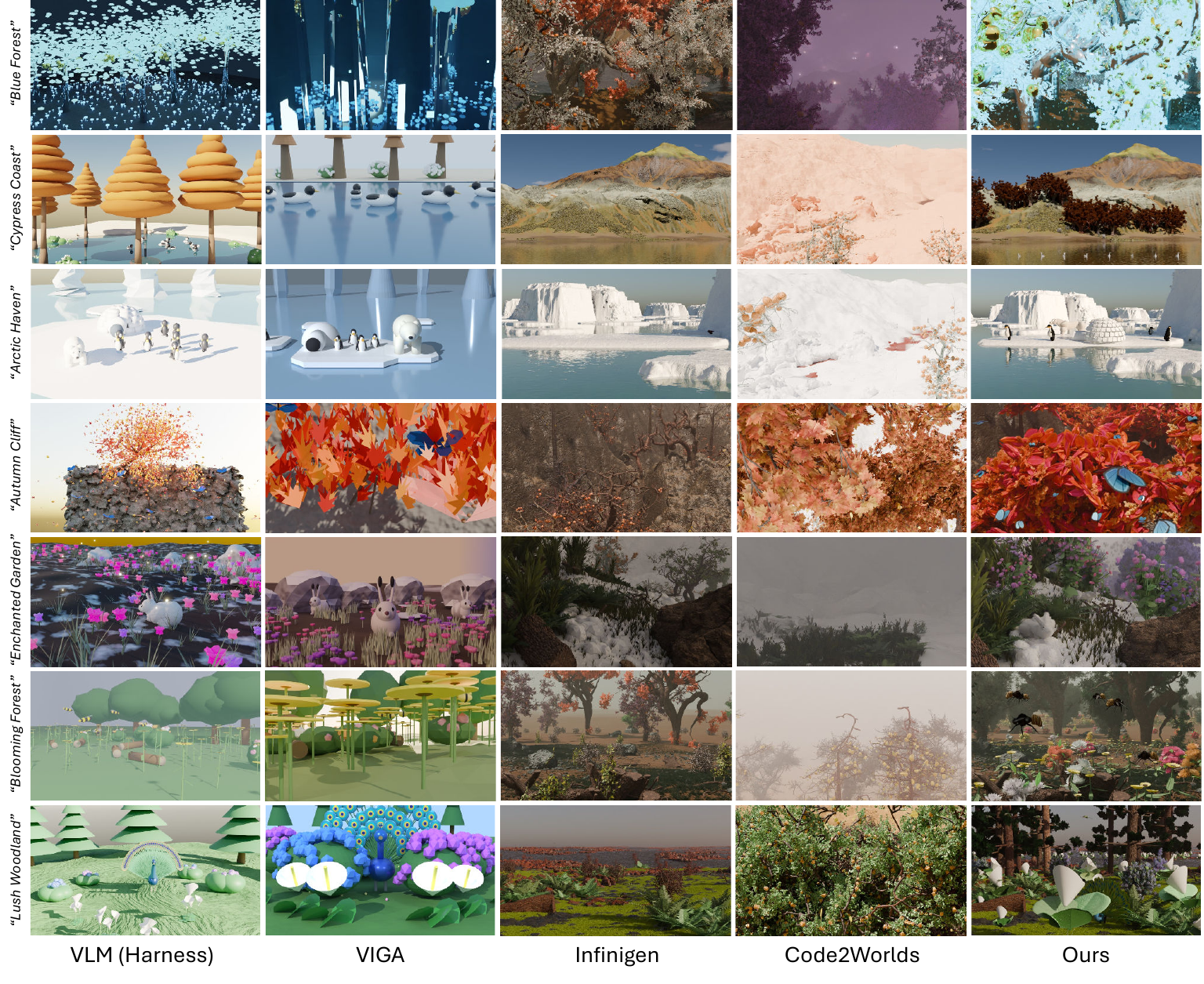}

    \caption{
        \textbf{Additional qualitative comparisons of scene-level generation.}
        More examples of 3D world generation across diverse
        environments under identical text prompts.
    }
    \label{fig:scene_app}
\end{figure}

We provide additional qualitative comparisons of scene-level
generation across diverse environments in Fig.~\ref{fig:scene_app}.
All methods are evaluated using identical text prompts under
the same experimental protocol described in Sec.~\ref{sec:exp_setup}.
These additional results illustrate differences in scene
composition, geometric detail, semantic alignment, and
visual fidelity across the compared methods.

\section{Additional Qualitative Comparison with GPT-6 Astra}
\label{app:astra_comparison}

\begin{figure}[H]
    \centering
    \includegraphics[
        width=\textwidth,
        height=0.88\textheight,
        keepaspectratio
    ]{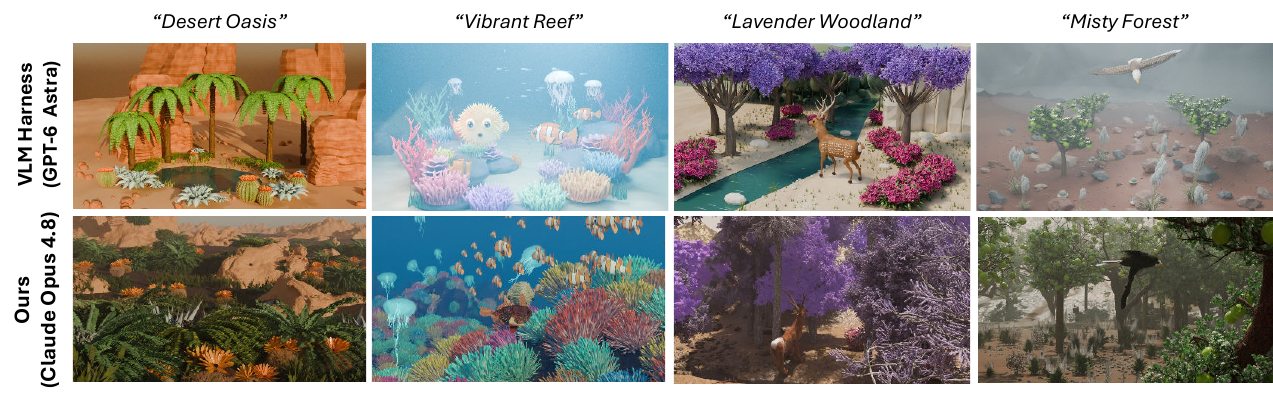}

    \caption{
        \textbf{Additional qualitative comparison with GPT-6 Astra.}
        We compare InfiniDream, implemented with Claude Opus 4.8,
        against GPT-6 Astra on representative examples under identical
        text prompts. The results provide additional evidence of
        InfiniDream's ability to generate visually rich and
        semantically coherent procedural assets and scenes.
    }
    \label{fig:astra_app}
\end{figure}

We provide an additional qualitative comparison with GPT-6 Astra
in Fig.~\ref{fig:astra_app}. This comparison is included as
supplementary evidence beyond the main benchmark suite.
Importantly, InfiniDream still uses Claude Opus 4.8 as the
underlying VLM, while GPT-6 Astra serves as an additional strong
proprietary baseline. Under identical text prompts, the
results highlight differences in visual fidelity, geometric
detail, material richness, and overall semantic composition.

\section{Additional Diversity Results}
\label{app:diversity_results}

\begin{figure}[H]
    \centering
    \includegraphics[
        width=\textwidth,
        height=0.88\textheight,
        keepaspectratio
    ]{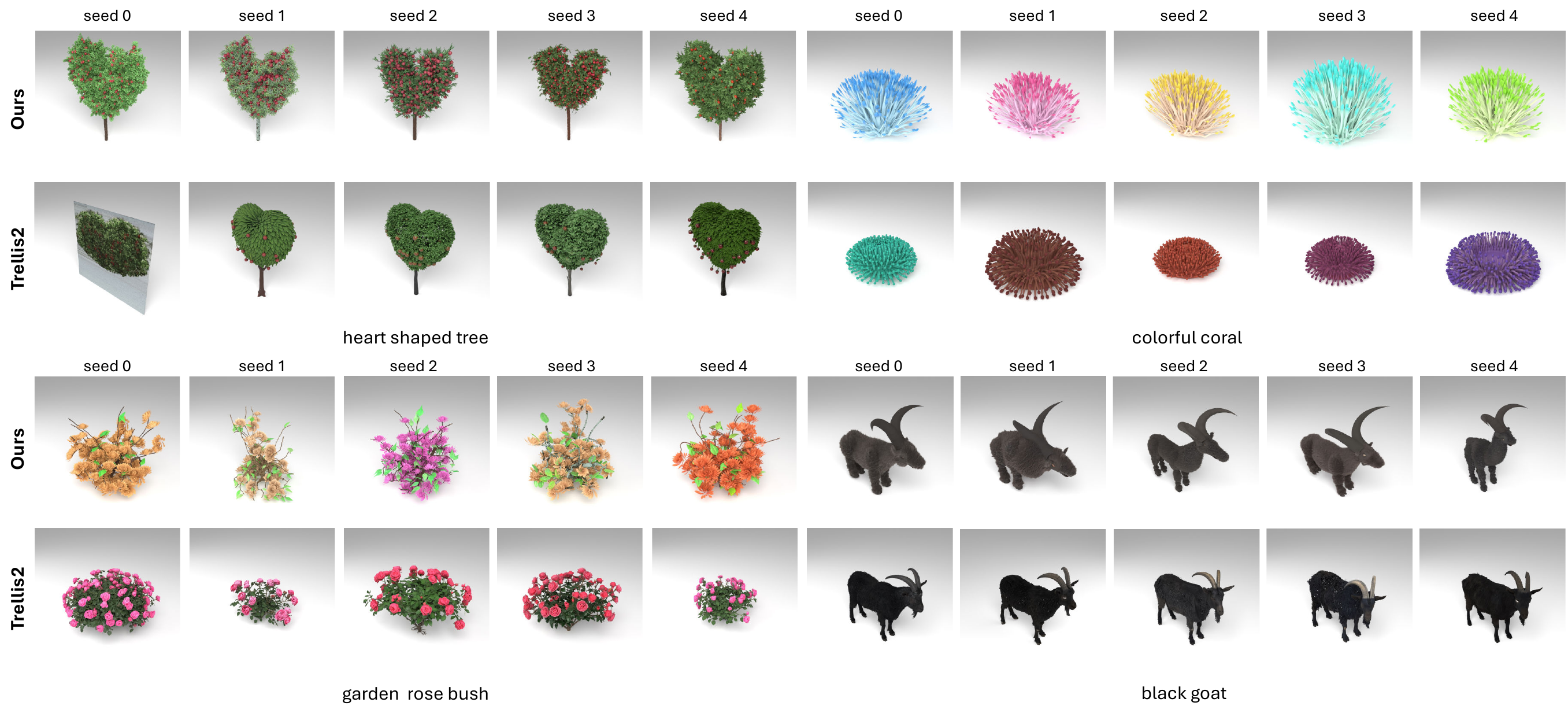}

    \caption{
        \textbf{Additional qualitative diversity comparisons
        with TRELLIS.2.}
        Five variations are generated for each text prompt.
        InfiniDream achieves procedural diversity by varying
        random seeds, whereas TRELLIS.2 uses independently
        generated reference images.
    }
    \label{fig:diversity_app}
\end{figure}

We provide additional qualitative diversity comparisons between
InfiniDream and TRELLIS.2 in Fig.~\ref{fig:diversity_app}.
For each text prompt, InfiniDream generates five procedural
variations by changing random seeds, while TRELLIS.2
generates five assets from independently generated reference
images. These examples further illustrate the diversity
of InfiniDream's procedural assets while preserving
semantic consistency and visual quality.

\section{Object-level Comparison with Code2Worlds}
\label{app:code2worlds_object}

\begin{figure}[!t]
    \centering

    \begin{minipage}[c]{0.55\linewidth}
        \centering
        \includegraphics[
            width=\linewidth,
            keepaspectratio
        ]{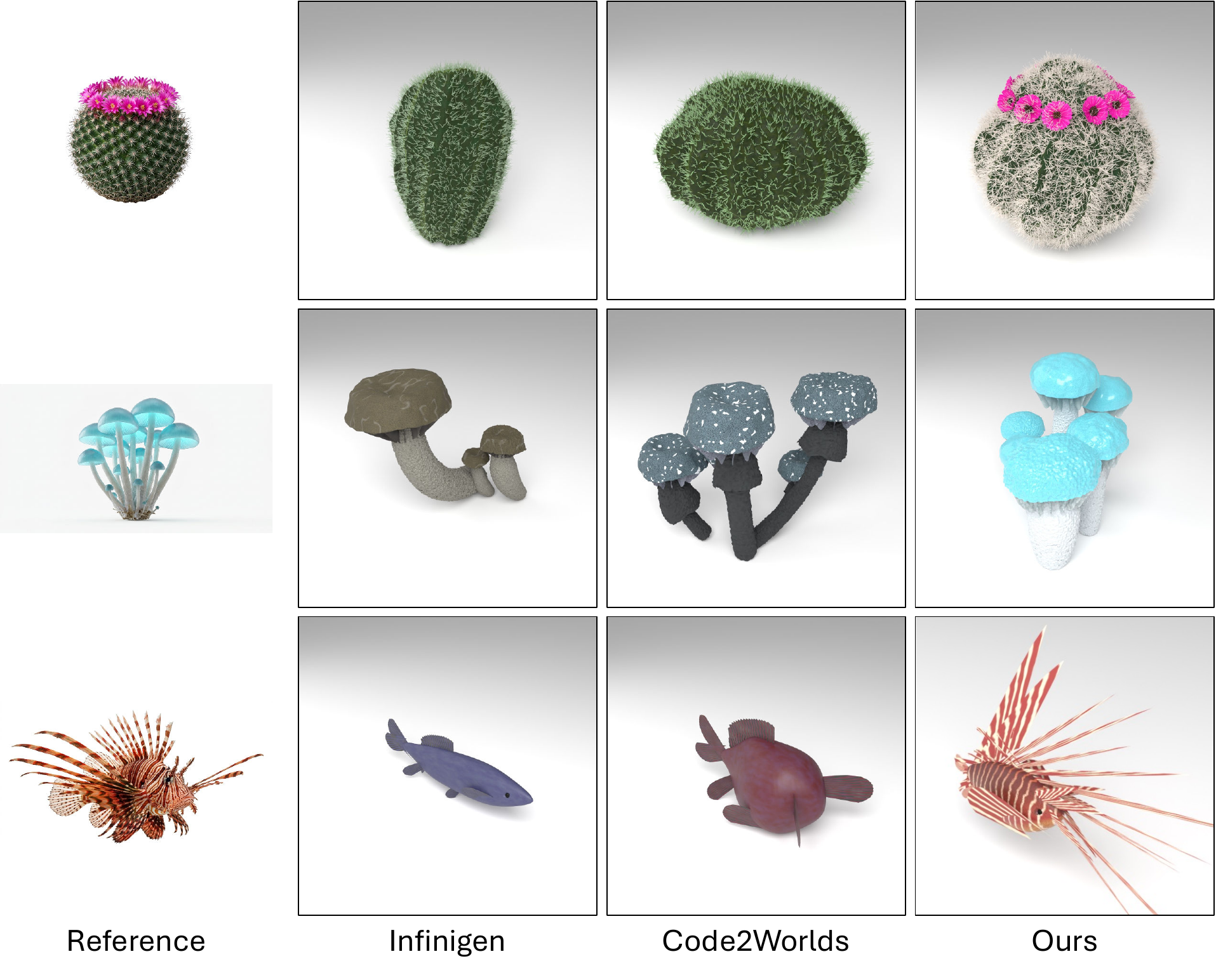}

        \caption{
            \textbf{Object-level comparison with Code2Worlds.}
        }
        \label{fig:code2worlds_object}
    \end{minipage}
    \hfill
    \begin{minipage}[c]{0.40\linewidth}
        \centering
        \captionof{table}{
            \textbf{Quantitative comparison.}
            Results on 29 supported tasks.
        }
        \label{tab:code2worlds_object}

        \vspace{4pt}
        \resizebox{0.9\linewidth}{!}{%
            \begin{tabular}{lcc}
                \toprule
                \textbf{Metric}
                & \textbf{InfiniDream}
                & \textbf{Code2Worlds} \\
                \midrule
                CLIP $\uparrow$
                & \textbf{30.24} & 22.56 \\
                SigLIP2 $\uparrow$
                & \textbf{0.153} & 0.080 \\
                ImageReward $\uparrow$
                & \textbf{-0.35} & -1.94 \\
                DINOv3 $\uparrow$
                & \textbf{0.583} & 0.281 \\
                \bottomrule
            \end{tabular}%
        }
    \end{minipage}

\end{figure}

We further compare InfiniDream with Code2Worlds
on the 29 out of 100 object-level tasks supported
by Code2Worlds. Due to its limited factory coverage,
we report this comparison separately from the main
benchmark.

As shown in Fig.~\ref{fig:code2worlds_object}
and Table~\ref{tab:code2worlds_object},
InfiniDream achieves higher scores across all four
metrics. Code2Worlds can exploit the parameter
space of existing factories, such as adjusting
the color of a fish, but cannot introduce novel
procedural features beyond predefined generators.
In contrast, InfiniDream extends existing procedural
code to generate new geometric structures and
fine-grained details, enabling more flexible
open-ended asset generation.

\section{Man-Made Asset Generation}
\label{app:manmade}

\begin{figure}[H]
    \centering
    \includegraphics[
        width=\textwidth,
        keepaspectratio
    ]{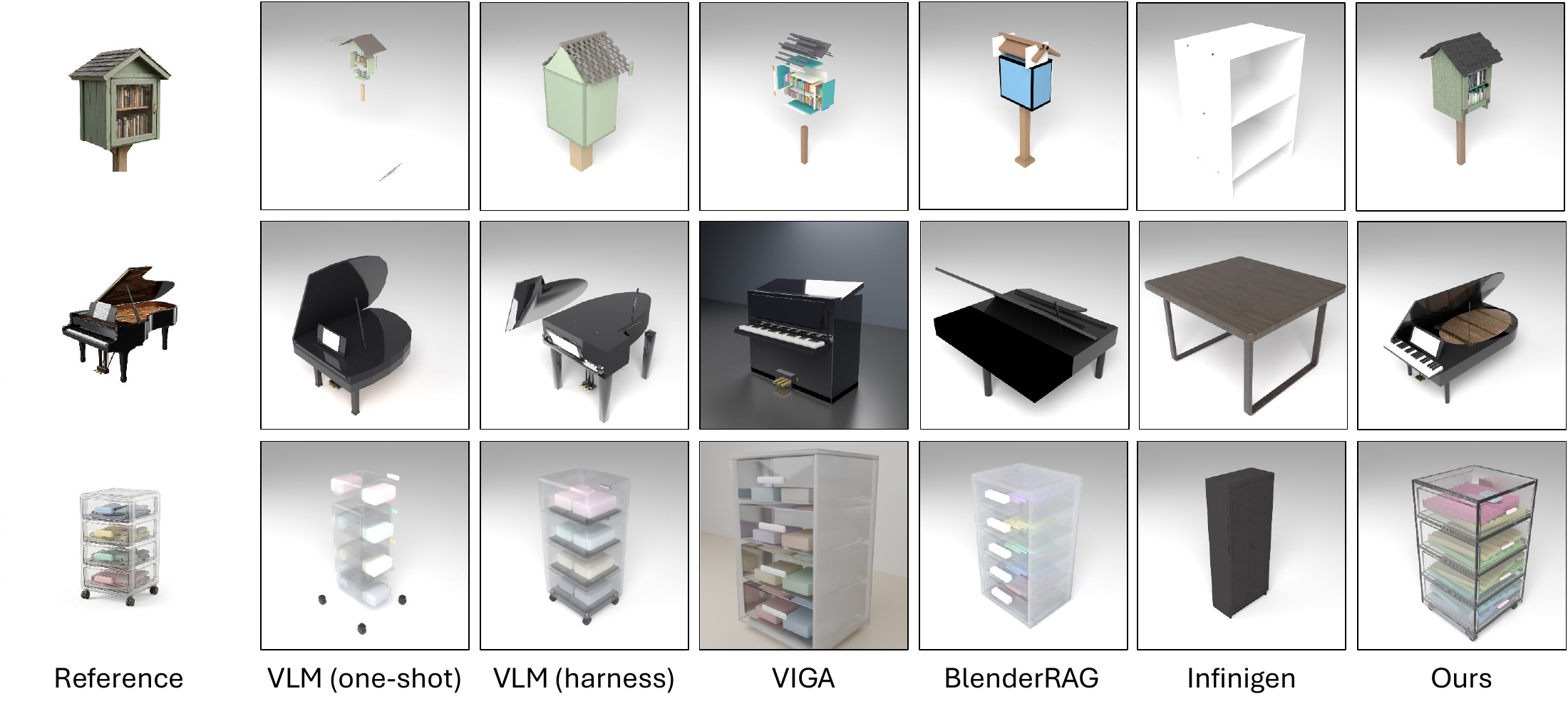}
    \caption{
        \textbf{Qualitative comparison of man-made asset generation.}
        InfiniDream produces structurally coherent
        man-made objects with rich geometric and material details.
    }
    \label{fig:manmade_app}
\end{figure}

Beyond natural assets, InfiniDream can retrieve and
extend factories for man-made objects. As shown in
Fig.~\ref{fig:manmade_app}, while VIGA performs well
on geometric objects, InfiniDream achieves comparable
structural quality with rich geometric and material
details, demonstrating its versatility across both
organic and man-made assets.

\section{Implementation Details of Symbolically Constrained Editing}
\label{app:sce}

\paragraph{Reference Synthesis and Factory Retrieval.}
We use Gemini 3 Pro Image to generate an isolated
reference object against a white background.
Claude Opus 4.8 retrieves the most relevant Infinigen
factory, with deterministic lexical matching as a
fallback.

\paragraph{Factory Brief Construction.}
Starting from the selected factory, we perform
AST-based import graph traversal with a maximum
depth of two. Asset-specific modules form the
Editable Surface, while shared modules remain
read-only Linked Infrastructure. Unrelated
dependencies are pruned, and a VLM summarizes
the remaining code into a cached Code Structure Map.

\paragraph{Constrained Code Editing.}
The Planner produces an Editing Plan specifying
Strategy, Locate, Edits, Rely-on, and Risks.
The Constructor can modify existing modules
(Tweak), promote protected modules (Promote),
or synthesize new procedural modules (Create).
Editing is restricted to the working directory,
preserving the original factory interface and
seed-dependent randomness.

\paragraph{Iterative Refinement.}
Modified modules are integrated into a symbolic-link
overlay of the original Infinigen codebase.
A vision-based Reviser guides targeted refinement, while execution failures
trigger traceback-based repair. The pipeline supports up to six rounds
with early stopping. A vision-based Selector
chooses the best successful candidate, whose
code snapshot is used to reconstruct the final
executable generator.

\section{Implementation Details of Deterministic Frustum-Guided Grid Placement}
\label{app:placement}

\paragraph{Candidate Generation.}
We discretize the Director's screen-space region into
a $9 \times 5$ grid ($17 \times 9$ for six or more instances).
Surface candidates are obtained through ray-scene
intersections, while airborne and submerged candidates
are sampled along camera rays using geometric depth
intervals ($\times 1.3$, up to 80\,m).
Larger objects are placed first, with deterministic
hashing ensuring reproducibility. The entire placement
process requires no additional VLM calls.

\paragraph{Constraints and Relaxation.}
Projected area is estimated from the proxy bounding box,
with a preferred minimum of 1.2\%, a relaxed minimum of
0.2\%, and an absolute maximum of 25\%.
Collision checks account for motion-dependent swept
envelopes, while surface placement additionally enforces
slope limits and four-corner footprint support.
When no valid candidate is found, constraints are
progressively relaxed and the search region is expanded.
Water-related constraints remain enforced, and all
relaxations are recorded in the resulting placement plan.

\section{Baseline Implementation Details}
\label{app:baseline}

\paragraph{Baseline Implementations.}
We adopt the official implementations of all baselines,
adding Claude Opus 4.8 API adapters where necessary.
The VLM (one-shot) and VLM (harness) baselines follow
3DCodeBench, with the latter supporting autonomous
code execution and iterative refinement.

\paragraph{SigLIP2 Score Computation.}
All SigLIP2 scores in this paper (Tables~1, 2, 4, 5, 7) are computed with
\texttt{google/siglip2-base-patch16-224} and reported as the cosine similarity
between L2-normalized image and text embeddings, recovered from the model's
image--text logits via $\cos = (\mathrm{logit} - b)/t$, where $t = 112.67$ and
$b = -16.77$ are SigLIP2's learned temperature and bias. Scores are averaged
over the four azimuth views at the object level (Tables~1, 5, 7), the primary
camera view at the scene level (Table~2), and the five variations per prompt
in the diversity evaluation (Table~4).

\paragraph{Object-level Baselines.}
All methods receive identical text prompts and reference images, except
BlenderRAG, which only supports text inputs. Each task has five Gemini-generated
reference images: \texttt{ref\_00} is the conditioning reference provided to the
generation methods, while \texttt{ref\_01}--\texttt{ref\_04} are held out
exclusively for evaluation (DINOv3) and the human study and are never shown to
any method; the conditioning reference is never used for scoring. All assets are
evaluated under identical camera, lighting, background, and rendering settings.
Failed generations are retained as blank renders rather than excluded.

\paragraph{Scene-level Baselines.}
All scene-level methods receive identical text prompts and follow their
respective official scene generation and camera selection pipelines. For
InfiniDream, the Scene Picker automatically retrieves the most relevant initial
coarse scene based on the input prompt from a library of 70 pre-generated
environments (14 biomes with five seeds each), which is then used as the basis
for subsequent procedural editing. The library was built before the benchmark
prompts were finalized and was not modified afterwards. For reproducibility, the
Scene Picker's selection is made once per task, and the chosen base scene and
seed are then fixed across all benchmark and ablation runs. The other baselines
do not use this library. Code2Worlds selects from its available procedural
factories even when no closely matching category exists.

\paragraph{Failure Handling.}
We resolve version- and environment-related execution
issues wherever possible to ensure that baselines
run as intended. We do not report a separate
scene-level executability metric.

\section{Ablation Study Details}
\label{app:ablation}

\paragraph{Object-level Ablation.}
We provide additional implementation details for the six
ablation settings:
(1) \textit{w/o Ref. Image} removes the reference image
from the generation pipeline;
(2) \textit{Random Factory} replaces factory retrieval
with random selection;
(3) \textit{w/o Planner} removes structured planning
and replaces the Editing Plan with empty content;
(4) \textit{Agent (Infinigen)} gives a Claude Opus 4.8
coding-agent harness access to the entire Infinigen
codebase and Bash tools, allowing autonomous code
generation, rendering, and refinement without SCE
constraints, for up to six rounds;
(5) \textit{w/o Refiner} removes fine-level refinement;
and (6) \textit{3 Rounds} limits the refinement budget
to three rounds, compared with six in the full method.

\paragraph{Scene-level Ablation.}
We evaluate four placement variants:
(1) \textit{w/o Grid Search} removes discrete semantic
planning and grid-based placement, allowing the
Director to predict continuous 3D coordinates directly;
(2) \textit{w/o Anchoring} replaces our seven anchoring
types with Infinigen's original grounding mechanism;
(3) \textit{w/o Semantic Target} removes semantic
surface constraints from grid search, allowing
placements on semantically inappropriate surfaces;
and (4) \textit{Agent (Placement)} provides a VLM
agent with our generated 3D assets, the text prompt,
and the same selected coarse scene, allowing
autonomous placement without our placement pipeline.

\paragraph{Contact Spread.}
For each object, we select the lowest 12\% of sampled points and cluster them
into four contact regions using k-means. The lowest point in each region is
selected, and its vertical gap to the underlying terrain is computed via ray
casting. Contact Spread is defined as the difference between the maximum and
minimum gaps, capturing uneven grounding independently of terrain elevation.

Both contact metrics are computed only for surface-supported objects (rooted,
upright, tilted, conform, and afloat; afloat is measured against the water
surface); airborne and submerged objects are excluded, while the Unreadable
Rate covers all placed objects. Ground-contact Error uses the absolute gap, so
floating and penetration both count by magnitude. Objects whose contact rays
find no supporting surface, or that failed to populate, are excluded. P95 is
the nearest-rank 95th percentile of per-object values pooled across all
surface-supported requested-add objects in the 20 scenes. For \textit{Agent
(Placement)}, anchoring types are inferred from the full-pipeline runs; objects
without a counterpart, and plant assets whose lowest vertices are foliage
rather than ground contacts, are excluded from its contact statistics.

\section{Detailed Capability Comparison}
\label{app:capability}

We provide a detailed capability comparison of
representative 3D generation methods in
Table~\ref{tab:capability}. The comparison covers
three asset-level and five scene-level capabilities,
highlighting InfiniDream's integration of open-ended
procedural asset generation with scalable world
construction.

\begin{table*}[t]
    \centering
    \caption{
        \textbf{Capability comparison of representative
        3D generation methods.}
        We compare three asset-level and five scene-level
        capabilities. ``Novel Proc. Assets'' denotes the
        creation of procedural generators beyond predefined
        factories, while ``Proc. Editing'' includes editable
        procedural representations. ``Placement Reasoning''
        refers to semantic spatial planning, whereas
        ``Semantic Anchoring'' requires semantic support-surface
        selection and geometry-aware contact alignment.
        \yes{} indicates explicit support, $\triangle$ partial
        support, and \no{} support not established under
        our criteria.
    }
    \label{tab:capability}

    \resizebox{\textwidth}{!}{%
    \begin{tabular}{lcccccccc}
        \toprule

        & \multicolumn{3}{c}{\textbf{Asset-level}}
        & \multicolumn{5}{c}{\textbf{Scene-level}} \\

        \cmidrule(lr){2-4}
        \cmidrule(lr){5-9}

        \textbf{Method}
        & \shortstack{\textbf{Novel Proc.}\\\textbf{Assets}}
        & \shortstack{\textbf{Seed}\\\textbf{Diversity}}
        & \shortstack{\textbf{Proc.}\\\textbf{Editing}}
        & \shortstack{\textbf{Text-to-}\\\textbf{Scene}}
        & \shortstack{\textbf{Placement}\\\textbf{Reasoning}}
        & \shortstack{\textbf{Semantic}\\\textbf{Anchoring}}
        & \shortstack{\textbf{Scalable}\\\textbf{Population}}
        & \textbf{Animation} \\

        \midrule

        VIGA~\citep{yin2026viga}
        & \yes & \no & \yes
        & \yes & \yes & \no & \no & \yes \\

        LL3M~\citep{lu2025ll3m}
        & \yes & \no & \yes
        & \no & \no & \no & \no & \no \\

        Infinigen~\citep{infinigen2023infinite}
        & \no & \yes & \yes
        & \no & \no & \no & \yes & \yes \\

        3D-GPT~\citep{sun20233dgpt}
        & \no & \yes & \yes
        & \yes & \yes & \no & \yes & \yes \\

        SceneX~\citep{zhou2025scenex}
        & \no & \yes & \yes
        & \yes & \yes & $\triangle$ & \yes & \yes \\

        Code2Worlds~\citep{zhang2026code2worlds}
        & \no & \yes & \yes
        & \yes & \yes & \no & \yes & \yes \\

        WorldClaw~\citep{guo2026worldclaw}
        & \no & \no & \no
        & \yes & \yes & \yes & \yes & \no \\

        WorldGen~\citep{wang2026worldgen}
        & \no & \no & \no
        & \yes & \yes & $\triangle$ & \yes & \no \\

        \midrule

        \textbf{InfiniDream}
        & \yes & \yes & \yes
        & \yes & \yes & \yes & \yes & \yes \\

        \bottomrule
    \end{tabular}%
    }
\end{table*}

\section{Human Evaluation Details}
\label{app:userstudy}

\begin{figure}[H]
    \centering
    \includegraphics[
        width=\textwidth,
        keepaspectratio
    ]{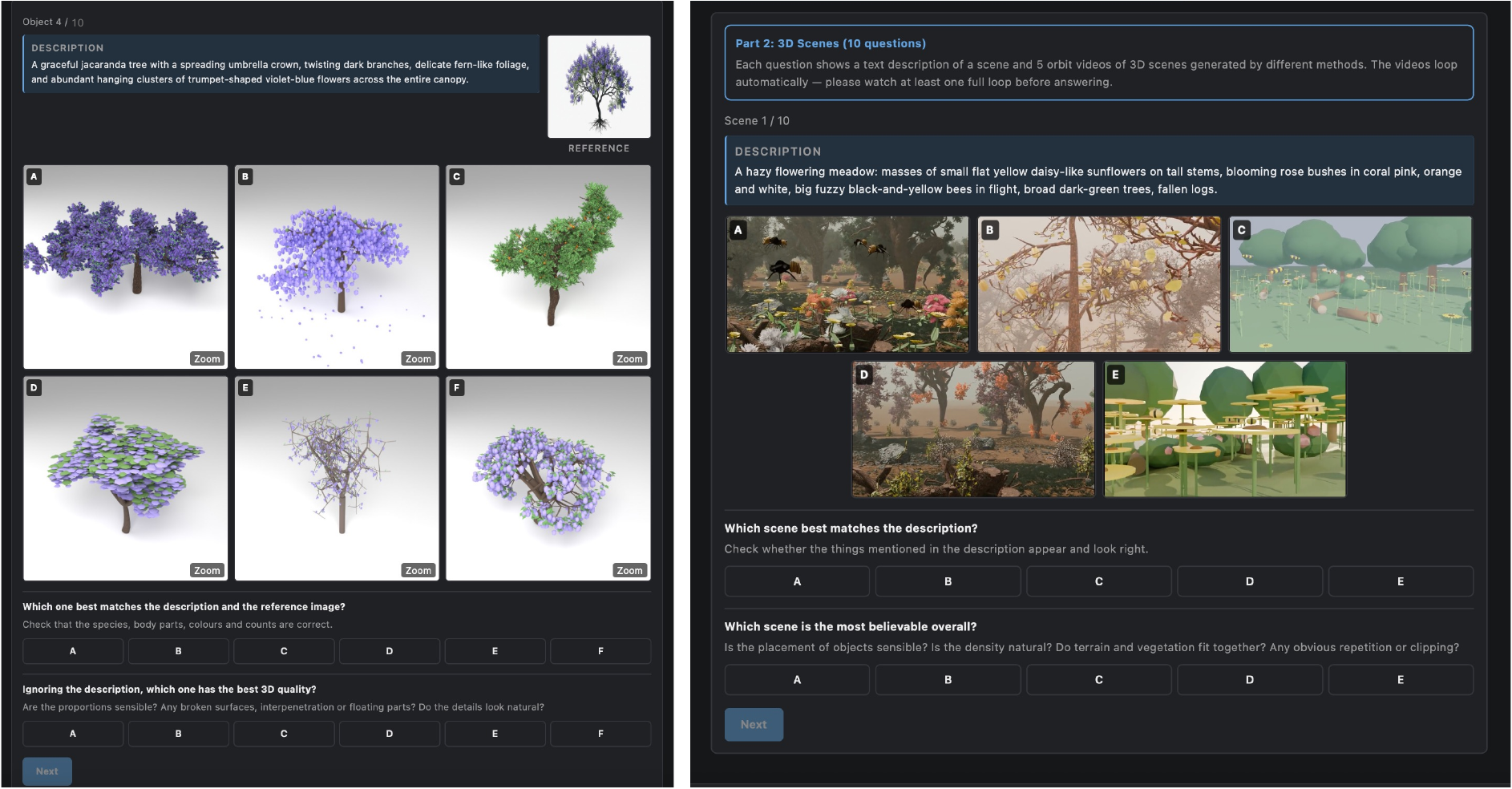}

    \caption{
        \textbf{Human evaluation interface.}
        Participants compare anonymized generation
        results and independently select their
        preferred method for semantic alignment
        and visual quality at both the object
        and scene levels.
    }
    \label{fig:userstudy}
\end{figure}

\paragraph{Study Setup.}
We conduct a forced-choice preference study on
10 object and 10 scene generation tasks.
For each object, participants are shown the text
prompt, a held-out reference image not provided to
any generation method, and front-view renders
($1024 \times 1024$) from six methods:
InfiniDream, BlenderRAG, VIGA, VLM (one-shot),
VLM (harness), and Infinigen.

For each scene, participants compare 5-second
orbit videos ($854 \times 480$) from five methods:
InfiniDream, VIGA, VLM (harness), Code2Worlds,
and Infinigen. All results are presented without
cropping or background normalization.

The object subset covers animals, marine life,
plants, and natural props, while the scene subset
spans seven distinct environment types.

\paragraph{Evaluation Protocol.}
As illustrated in Fig.~\ref{fig:userstudy},
all methods are anonymized, and both task order
and method positions are randomized independently
for each participant. Each task contains two
forced-choice questions evaluating semantic
alignment and visual quality.

For objects, participants select the result that
best matches the description and reference image,
and the result with the highest 3D quality.
For scenes, participants evaluate text alignment
and overall scene realism, considering object
placement, naturalness, and visual artifacts.

Each participant completes 40 judgments across
both levels. We recruit 30 participants,
yielding 20 judgments per level per participant.